\documentclass[runningheads]{llncs}
\usepackage[latin9]{inputenc}
\usepackage{float}
\usepackage{multirow}
\usepackage{amsmath}
\usepackage{amssymb}
\usepackage{dirtytalk}
\usepackage{graphicx}

\makeatletter

\providecommand{\tabularnewline}{\\}
\floatstyle{ruled}
\newfloat{algorithm}{tbp}{loa}
\providecommand{\algorithmname}{Algorithm}
\floatname{algorithm}{\protect\algorithmname}

\usepackage{comment}
\usepackage{color}

\usepackage{algorithm}
\usepackage{algorithmic}

\@ifundefined{showcaptionsetup}{}{%
 \PassOptionsToPackage{caption=false}{subfig}}
\usepackage{subfig}
\makeatother

\begin{document}
\global\long\def\ECCVSubNumber{3177}%

\title{CATCH: Context-based Meta Reinforcement Learning for Transferrable
Architecture Search}
\titlerunning{CATCH}
\authorrunning{X. Chen and Y. Duan et al.}
\author{Xin Chen\thanks{Equal contribution.}\inst{1}, Yawen Duan$^{\star}$\inst{1}, Zewei Chen\inst{2}, Hang Xu\inst{2}, Zihao Chen\inst{2},\\ Xiaodan Liang\inst{3}, Tong Zhang\thanks{Correspondence to: tongzhang@tongzhang-ml.org}\inst{4}, Zhenguo Li\inst{2}}
\institute{The University of Hong Kong \and Huawei Noah's Ark Lab \and Sun Yat-sen University \and The Hong Kong University of Science and Technology}
\maketitle

\begin{abstract}
Neural Architecture Search (NAS) achieved many breakthroughs in recent years. In spite of its remarkable progress, many algorithms are restricted
to particular search spaces. They also lack efficient mechanisms to
reuse knowledge when confronting multiple tasks. These challenges
preclude their applicability, and motivate our proposal of CATCH,
a novel Context-bAsed meTa reinforcement learning (RL) algorithm for transferrable
arChitecture searcH. The combination of meta-learning and RL allows CATCH to efficiently adapt to new tasks while being agnostic to search spaces. CATCH utilizes a probabilistic encoder to encode task properties into latent context variables, which then guide CATCH\textquoteright s controller to quickly \say{catch} top-performing networks. The contexts also assist a network
evaluator in filtering inferior candidates and speed up learning. Extensive
experiments demonstrate CATCH\textquoteright s universality and search
efficiency over many other widely-recognized algorithms. It is also
capable of handling cross-domain architecture search as competitive
networks on ImageNet, COCO, and Cityscapes are identified. This is
the first work to our knowledge that proposes an efficient transferrable
NAS solution while maintaining robustness across various settings.

\end{abstract}

\section{Introduction}

The emergence of many high-performance neural networks has been one
of the pivotal forces pushing forward the progress of deep learning
research and production. Recently, many neural networks discovered
by Neural Architecture Search (NAS) methods have surpassed manually
designed ones on a variety of domains including image classification  \cite{tan2019efficientnet,zoph2018learning}, object detection \cite{zoph2018learning}, semantic segmentation \cite{chen2018searching}, and recommendation systems \cite{liu2020autofis}.
Many potential applications of practical interests are calling for
solutions that can (1) efficiently handle a myriad of tasks, (2) be
widely applicable to different search spaces, and (3) maintain their
levels of competency across various settings. We believe these are
important yet somewhat neglected aspects in the past research, and
a transformative NAS algorithm should be able to respond to these
needs to make a real influence.

\begin{figure}[t]
\begin{centering}
\includegraphics[width=1\columnwidth]{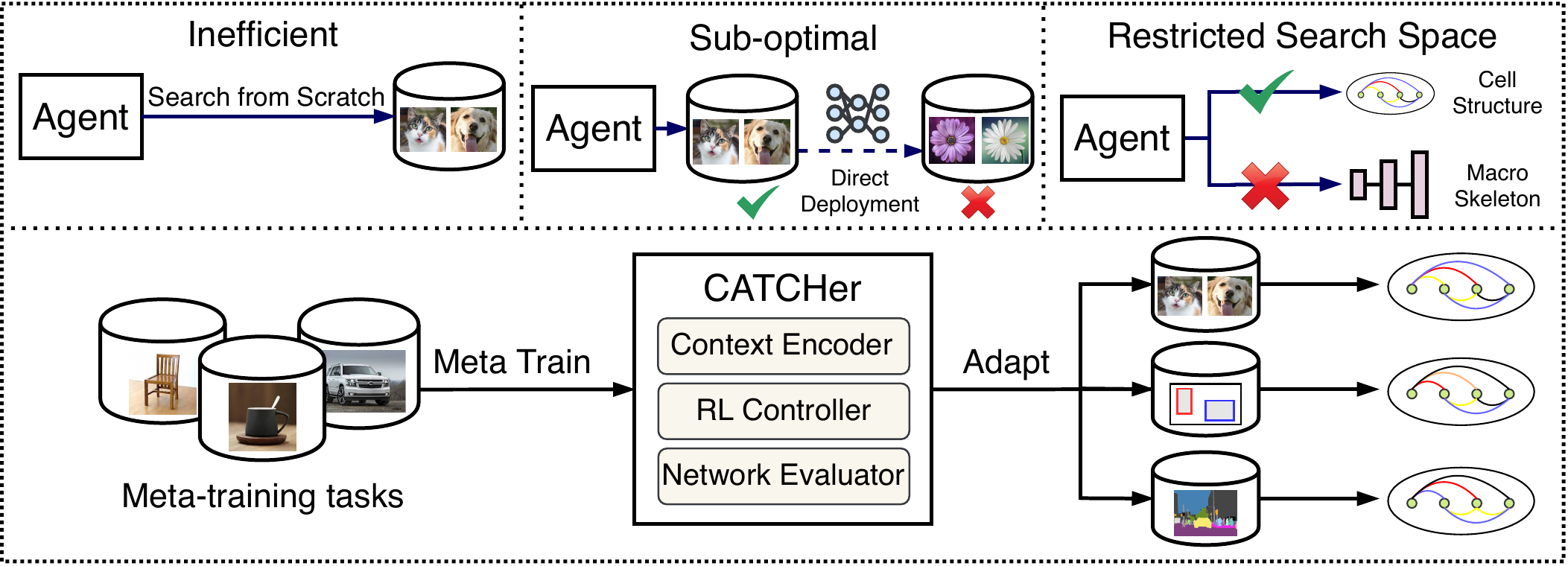}
\par\end{centering}

\caption{Upper: drawbacks of current NAS schemes. Lower: the overall framework
of CATCH. Our search agent, CATCHer, consists of three core
components: context encoder, RL controller and network evaluator.
CATCHer first goes through the meta-training phase to learn an initial
search policy, then it adapts to target tasks efficiently.}

\label{overall}

\end{figure}

Many algorithms \cite{liu2018darts,pham2018efficient} have been proposed
to improve the efficiency of NAS. However, they lack mechanisms to
seek and preserve information that can be meaningfully reused. Hence,
these algorithms can only repeatedly and inefficiently search from
scratch when encountering new tasks. To tackle this problem, a rising
direction of NAS attempts to create efficient transferrable algorithms.
Several works \cite{kim2018auto,pasunuru2019continual} try to search
for architectures that perform well across tasks, but the solutions
may not be optimal on the target tasks, especially when the target task distributions
are distant from the training task distributions. Some recent works \cite{Lian2020Towards,elsken2019meta}
use meta-learning \cite{finn2017model,li2017meta} for one-shot NAS instead. With recent critiques \cite{Yang2020NAS,li2019random}
pointing out some one-shot solutions\textquoteright{} dependence on
particular search spaces and sensitivity to hyperparameters, many
concerns arise on the practicality of these meta NAS works based on
one-shot methods. To avoid ambiguity, throughout this paper, \emph{tasks} are defined as problems that share the same action space, but differ in reward functions. In NAS, the change of either the dataset or domain (e.g. from classification to detection) alters the underlying reward function, and thus can be treated as different tasks.

Striking a balance between universality and efficiency is hard.
Solving the universality problem needs a policy to disentangle from
specifics of search spaces, which uproots an important foundation
of many efficient algorithms. The aim to improve efficiency on multiple
tasks naturally links us to a transfer/meta learning paradigm. Meta
Reinforcement Learning (RL) \cite{rakelly2019efficient,lan2019meta}
offers a solution to achieving both efficiency and universality,
which largely inspired our proposal of CATCH, a novel context-guided
meta reinforcement learning framework that is both search space-agnostic
and swiftly adaptive to new tasks. 

The search agent in our framework, namely CATCHer, acts as the decision-maker
to quickly \say{catch} top-performing
networks on a task. As is shown in Figure \ref{overall}, it is first
trained on a set of meta-training tasks then deployed to target tasks
for fast adaptation. CATCHer leverages three core components: context
encoder, RL controller, and network evaluator. The context encoder
adopts an amortized variational inference approach \cite{alemi2016deep,rakelly2019efficient,kingma2013auto}
to encode task properties into latent context variables that guide
the controller and evaluator. The RL controller makes sequential decisions
to generate candidate networks in a stochastic manner. The network
evaluator predicts the performance of the candidate networks and decides
which nets are valuable for training. All three components are optimized
in an end-to-end manner.

We test the method's universality and adaptation efficiency on two
fundamentally different search spaces: cell-based search space \cite{dong2020bench}
and Residual block-based \cite{he2016deep,yao2019sm} search space.
The former focuses on cell structure design, while the latter targets
macro skeleton search. With NAS-Bench-201 \cite{dong2020bench}, we
can compare CATCH fairly with other algorithms by eliminating performance
fluctuations rising from different search spaces and training settings.
Our experiments demonstrate CATCH's superiority over various other
works, including R-EA \cite{real2019regularized} and DARTS \cite{liu2018darts}.
On Residual block-based search space, we use image classification tasks on sub-datasets of ImageNet \cite{deng2009imagenet}
as meta-training tasks, and then adapt the CATCHer to target tasks, such as image classification on full ImageNet, object detection on COCO \cite{lin2014microsoft}, and semantic segmentation on Cityscapes \cite{Cordts2016Cityscapes}.
CATCH discovered networks on these tasks with competitive performance
and inference latency. Our results demonstrated CATCH\textquoteright s
robustness across various settings, easing previously raised concerns
of NAS algorithms' sensitivity to search space, random seeds, and
tendencies to overfit to only one or two reported tasks.

Our key contribution is the first attempt to design an efficient and
universal transferrable NAS framework. It swiftly handles various
tasks through fast adaptation, and robustly maintains competitive
performance across different settings. Our work brings along new perspectives
on solving NAS problems, including using amortized variational inference
to generate task characteristics that inform network designs. It also
demonstrates the possibility of creating efficient sample-based NAS
solutions that are comparable with widely-recognized one-shot methods.
With competitive networks identified across classification, detection,
and segmentation domains, it further opens the investigation on the
feasibility of cross-domain architecture search.

\section{Related Work}

NAS is an algorithmic approach to design neural networks through searching
over candidate architectures. Many harness the power of Reinforcement
Learning (RL) \cite{zoph2016neural}, Bayesian Optimization \cite{bergstra2013making,bergstra2011algorithms},
Evolutionary Algorithm \cite{elsken2018efficient,real2019aging}, and
Monte Carlo Tree Search \cite{negrinho2017deeparchitect,wistuba2017finding}.
The field gradually gains its tractions with the emergence of highly-efficient
algorithms \cite{liu2018darts,pham2018efficient,real2019aging} and
architectures \cite{real2019regularized,tan2019efficientnet} with
remarkable performance.

Our method is inspired by PEARL \cite{rakelly2019efficient}, a recent
work in context-based meta reinforcement learning, which captures
knowledge about a task with probabilistic latent contexts. The knowledge
is then leveraged for informed policy training. There are
a few key challenges in efficiently applying it to NAS: (1) PEARL models the latent context embeddings of RL tasks as distributions over Markov Decision Processes (MDP), but it is less clear how a task in NAS can be meaningfully encoded. (2) RL is notoriously famous for its sample inefficiency,
but it is extremely expensive to obtain reward signals on NAS. We
address these challenges by (1) proposing the use of network-reward
pairs to represent a task, (2) introducing meta-training tasks that
can be cheaply evaluated to obtain more data for learning, and including
a network evaluator that acts like Q-learning agents to speed up learning.

Previous works also explored the possibility of using meta-learning
for NAS. Some \cite{kim2018auto,pasunuru2019continual} aimed to identify
a single architecture that simultaneously works well on all considered
tasks. These solutions may not be scalable when confronting
a large pool of target tasks. An early work \cite{wong2018transfer}
aimed to learn a general policy across tasks. However, it generates
task embeddings from images, which may fail at datasets with the
same images, and is unable to differentiate among classification,
detection, and segmentation tasks on the same dataset. A few recent
papers \cite{Lian2020Towards,elsken2019meta} combined gradient-based
meta-learning with DARTS, but the algorithms are only applicable to
search spaces compatible with DARTS. Additionally, none of the above
proposals reported their performance on large-scale tasks like ImageNet
full dataset. This leaves questions on these proposals' generalizability
and adaptation efficiency on more challenging datasets, where scientists
expect meta-NAS algorithms should have an edge over typical NAS methods.
CATCH is the first NAS algorithm to our knowledge that deploys meta-learning
while maintaining universality, robustness across different search
spaces, and capability to handle large-scale tasks.

\section{CATCH Framework\label{sec:Methods}}

In NAS, the change of dataset (e.g. CIFAR-10 vs. ImageNet) or domain (e.g. image classification vs. object detection) essentially indicates the shift of underlying reward distribution. The goal of a cross-task transfer algorithm is hence to quickly identify the best actions under the changed reward dynamics. To handle this challenge, the CATCH framework consists of two phases: a meta-training phase and an adaptation phase, as is presented in Algorithm 1. In the meta-training phase, we train the CATCHer on a pool of meta-training tasks that can be cheaply evaluated. A key goal of this phase is to present the context encoder with sufficiently diversified tasks, and encourage it to consistently encode meaningful information for different tasks. Meanwhile, both the controller and the evaluator may gain a good initialization for adaptation. In the adaptation phase, the meta-trained CATCHer then learns to find networks on the target task efficiently through the guidance of the latent context encoding. 

\begin{figure}[t]
\begin{centering}
\includegraphics[width=1\columnwidth]{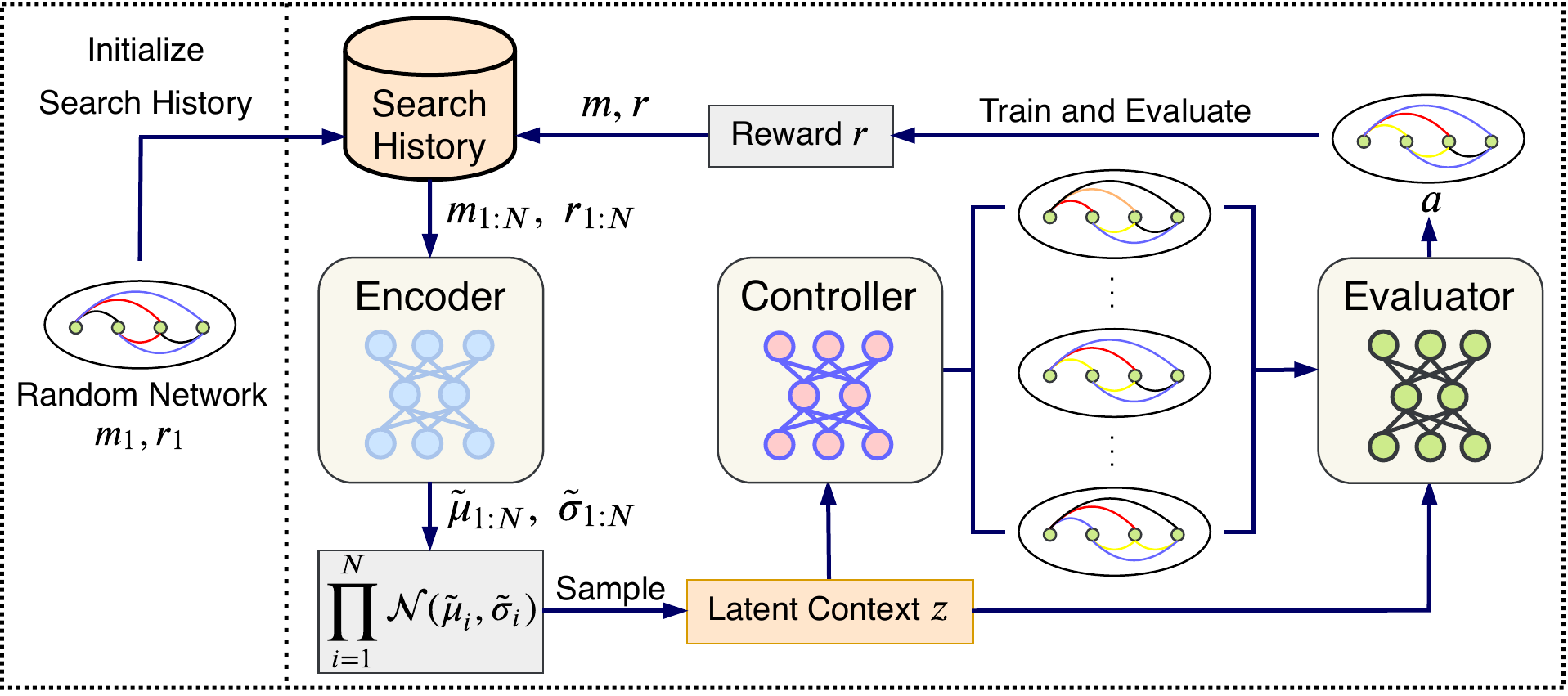}
\par\end{centering}

\caption{The search procedure of CATCH on a given task. The procedure starts
from initializing the search history by storing a randomly selected
network $m$ and its reward $r$. The encoder applies amortized variational
inference approach to generate latent context encoding $\boldsymbol{z}$
by encoding network-reward pairs from the search history. The controller
then generates candidate networks for the evaluator to choose the
most promising ones to train and evaluate. Newly selected networks
and their rewards will be stored in the search history. The loop continues
after the three components are optimized.}

\label{procedure}

\end{figure}

We show the search procedure on any single task in Figure  \ref{procedure},
which corresponds to line 3-13 of Algorithm  \ref{CATCH}.

\subsection{Context Encoding}

The use of latent context encoding is a crucial part of CATCH. The question
is what information about the task is reliable to construct such
latent contexts. Directly extracting feature maps of images of the
dataset is an intuitive solution. However, for the same dataset, the
best network configurations to perform different tasks like object
detection and semantic segmentation may differ a lot. Hence, simply
extracting information directly from images may not be a viable approach.

We instead believe that the task-specific contextual knowledge
can be mined from the search history (i.e. sets of network-reward
pairs). If the same group of networks have similar relative strengths on two tasks, it might mean these tasks are \say{close} to each other. It is also helpful to break the barriers for
cross-task architecture search, since the network-reward pair of information
is universal across tasks.

Before searching on a task, we randomly form a few networks $m$
and evaluate their performance $r$ to initialize the search history.
The retrieved network-reward pairs are stored in the search history
for its initialization. To start the search, we sample a number of
network-reward pairs $\{(m,r)_{i}\}_{1}^{N}$ (denoted by $\boldsymbol{c}_{1:N}$
for simplicity) from the search history, which will be fed into the
encoder to generate a latent context vector $\boldsymbol{z}$ representing
the salient knowledge about the task.

We model the latent context encoding process in a probabilistic manner,
because it allows the context encoder to model a distribution over
tasks and conduct exploration via posterior sampling. Following the
amortized variational inference approach used in \cite{rakelly2019efficient,alemi2016deep,kingma2013auto},
we aim to estimate the posterior $p(\boldsymbol{z}|\boldsymbol{c}_{1:N})$
with the encoder $q_{\phi}(\boldsymbol{z}|\boldsymbol{c}_{1:N})$,
parametrized by $\phi $. We assume the prior $p(\boldsymbol{z})$ is
a unit multivariate Gaussian distribution with diagonal covariance
matrix $\mathcal{N}(\boldsymbol{0},diag(\boldsymbol{1}))$, and hence,
the posterior $p(\boldsymbol{z}|\boldsymbol{c})$ conditioning on $\boldsymbol{c}$
is Gaussian. Since the network-reward pairs $\boldsymbol{c}_{1:N}$
are independent on a task, we could factor $q_{\phi}(\boldsymbol{z}|\boldsymbol{c}_{1:N})$
into the product of Gaussian factors conditioning on each piece of
contexts $\boldsymbol{c}_{i}$,
\begin{equation}
q_{\phi}(\boldsymbol{z}|\boldsymbol{c}_{1:N})\propto\prod_{i=1}^{N}\mathcal{N}(f_{\phi}^{\tilde{\mu}}(\boldsymbol{c}_{i}),diag(f_{\phi}^{\tilde{\sigma}}(\boldsymbol{c}_{i})),
\end{equation}
where $f_{\phi}$ is an inference network parametrized by $\phi$,
which predicts the mean $\tilde{\boldsymbol{\mu}}_i$ and the standard deviation $\tilde{\boldsymbol{\sigma}}_i$
of $q_{\phi}(\boldsymbol{z}|\boldsymbol{c}_{i})$ as a function of $\boldsymbol{c}_{i}$ to approximate Gaussian $p(\boldsymbol{z}|\boldsymbol{c}_i)$.

During the forward pass, the encoder network $f_\phi$ outputs $\tilde{\boldsymbol{\mu}}_i$, $\tilde{\boldsymbol{\sigma}}_i$ of the Gaussian posterior $q_{\phi}(\boldsymbol{z}|\boldsymbol{c}_{i})$ conditioning on each context, then we take their product $q_{\phi}(\boldsymbol{z}|\boldsymbol{c}_{1:N})$. Each context $\boldsymbol{c}_i$ is $(m,r)_i$, where $r$ is normalized among $\{r\}_{1:N}$ to reflect the relative advantage of each network. All the network-reward pairs in the search history are utilized. We then sample $\boldsymbol{z}$ from $q_{\phi}(\boldsymbol{z}|\boldsymbol{c}_{1:N})$. Further implementation details can be found in the Appendix.

\subsection{Network Sampling}

The generation of a network can be treated as a decision-making problem,
where each of the RL controller's actions determines one attribute
of the resulting architecture. The attribute can be an operation type
to form a certain edge in a cell-based search (e.g. skip-connect, convolution
operations, etc.), or the shape of a network in a macro-skeleton search
(e.g. width, depth, etc.). Both ways are explored in our work.

A network, denoted by $m$, is represented as a list of actions $[a_{1},a_{2},...,a_{L}]$
taken by the controller in a sequential manner. At each time step $l$,
the controller makes a decision $a_l$ according to its policy $\pi_{\theta_{c}}$,
parametrized by $\theta_{c}$. The controller policy takes $\boldsymbol{z}$
and the previous actions $[a_{1}...a_{l-1},\boldsymbol{0},...,\boldsymbol{0}]$ as inputs, and outputs
the probability distribution of choosing a certain action $\pi_{\theta_{c}}(a^{l}|[a_{1}...a_{l-1},\boldsymbol{0},...,\boldsymbol{0}],\boldsymbol{z})$,
where the actions will be sampled accordingly. $\boldsymbol{z}$ is
the latent context vector generated by the encoder, and $[a_{1}...a_{l-1},\boldsymbol{0},...,\boldsymbol{0}]$
is a collection of one-hot vectors indicating all the actions taken
so far at $l$-th timestep, leaving untaken actions $[a_{l},...,a_{L}]$ as zero vectors. The reward for each action is the normalized performance score of the network. The controller samples $M$ networks stochastically as candidates for the network evaluator.

\subsection{Network Scoring and Evaluation}

Since the candidate networks are sampled stochastically by the controller,
it is almost inevitable that some inferior models will be generated.
We set up a filtering mechanism, namely network evaluator, which acts
like a Q-learning agent that predicts the actual performance of each
network, and selects the top one for training. The predicted value
is not necessarily an accurate prediction of the training performance,
but should be able to provide a ranking among candidate models roughly
similar to their true performance.

The evaluator $f_{\theta_{e}}(m,\boldsymbol{z})$ is parameterized by $\theta_{e}$.
It takes $M $ tuples of network-context pairs $(m,\boldsymbol{z})$
as inputs, and outputs the predicted performance of input architectures.
The network with the highest predicted performance score will be trained
to obtain the true reward $r$. The network-context-reward tuple $(m,\boldsymbol{z},r)$
is then stored in the evaluator's local memory for future gradient updates.

\begin{algorithm}[t]
\begin{algorithmic}[1]
\global\long\def\algorithmicrequire{\textbf{Inputs:}}%
\REQUIRE $\{\mathcal{T}_{meta}\}$ (meta-training task pool), $\{\mathcal{T}_{target}\}$ (target task pool), $N_{meta}$ (\# of meta epochs), $N_{search}$ (\# of
search epochs), $C $ (\# of contexts to sample), $M$ (\# of models to
sample)

\textbf{Meta-training Phase:}

\FOR{$N_{meta}$ meta epochs}

\STATE Select meta-training task $\mathcal{T}$ from $\{\mathcal{T}_{meta}\}$

\STATE Initialize SearchHistory

\FOR{$n=1$ to $N_{search}$}

\STATE $\{(m,r)_{i}\}_{1}^{C}=$ SearchHistory.sample\_contexts($C $)
\STATE$\boldsymbol{z}=\ $Encoder.encode($\{(m,r)_{i}\}_{1}^{C}$)
\STATE $\{m\}_{1}^{M}\leftarrow\ $Controller.sample\_networks($\boldsymbol{z}$, $M $)
\STATE$m'\leftarrow\ $Evaluator.choose\_best($\{m_{j}\}_{1}^{M}$, $\boldsymbol{z}$)
\STATE $r\leftarrow\ $train\_and\_evaluate($m',\mathcal{T}$)
\STATE SearchHistory.add($(m',\boldsymbol{z},r)$)
\STATE Encoder, Controller, Evaluator optimization
\ENDFOR\ENDFOR

\textbf{Adaptation Phase:}

\STATE Select target task $\mathcal{T}$ from $\{\mathcal{T}_{target}\}$
\STATE\textbf{Repeat} Line 3-13
\STATE BestModel $\leftarrow$ SearchHistory.best\_model()\RETURN
BestModel\end{algorithmic}
\caption{Context-based Meta Architecture Search (CATCH)}
\label{CATCH}
\end{algorithm}

\subsection{Optimization of CATCHer}

To optimize the controller policy, we maximize the expected reward for the task it is performed
on. The controller is trained using Proximal Policy Optimization (PPO) \cite{schulman2017proximal}
with a clipped surrogate objective $\mathcal{L}_{c}$. 

To optimize the evaluator, we deploy Prioritized Experience Replay (PER) \cite{schaul2015prioritized},
a Deep Q-learning \cite{mnih2013playing} optimization technique. During
the update, it prompts the evaluator to prioritize sampling entries
that it makes the most mistakes on, and thus improves sample efficiency.
The loss of the evaluator $\mathcal{L}_{e}$ is the Huber loss \cite{huber1992robust}
between the evaluator's prediction $\tilde{r}$ and the normalized true performance score. Further details of $\mathcal{L}_{c}$ and $\mathcal{L}_{e}$
can be found in the Appendix.

To optimize the encoder, we take $\mathcal{L}_{c}$
and $\mathcal{L}_{e}$ as part of the objective. The resulting variational lower bound for each task $\mathcal{T}$
is
\begin{equation}
\mathcal{L}=\mathbb{E}_{\boldsymbol{z}\sim q_{\phi}(\boldsymbol{z}|c^{\mathcal{T}})}[\mathcal{L}_{c}+\mathcal{L}_{e}+\beta D_{KL}(q_{\phi}(\boldsymbol{z}|\boldsymbol{c}^{\mathcal{T}})||p(\boldsymbol{z}))],
\label{loss}
\end{equation}

where $D_{KL}$ serves as an approximation to a variational information
bottleneck that constrains the mutual information between $\boldsymbol{z}$
and $\boldsymbol{c}$, as is shown in \cite{alemi2016deep,rakelly2019efficient}.
This information bottleneck acts as a regularizer to avoid overfitting
to training tasks. $\beta$ is the weight of $D_{KL}$
in the objective, and $p(\boldsymbol{z})$ is a unit
Gaussian prior. Since (1) the latent context $\boldsymbol{z}$ serves as input to both controller and evaluator, and (2) $q_{\phi}(\boldsymbol{z}|\boldsymbol{c})$ and $p(\boldsymbol{z})$ are Gaussian, with $D_{KL}$ computed using their mean and variance, gradient of Eq. \ref{loss} can be back-propagated end-to-end to the encoder with the reparameterization trick.

\section{Experiments\label{sec:Experiments}}

\begin{figure}[t]

\begin{centering}
\subfloat[CIFAR-10]{\includegraphics[width=0.33\columnwidth]{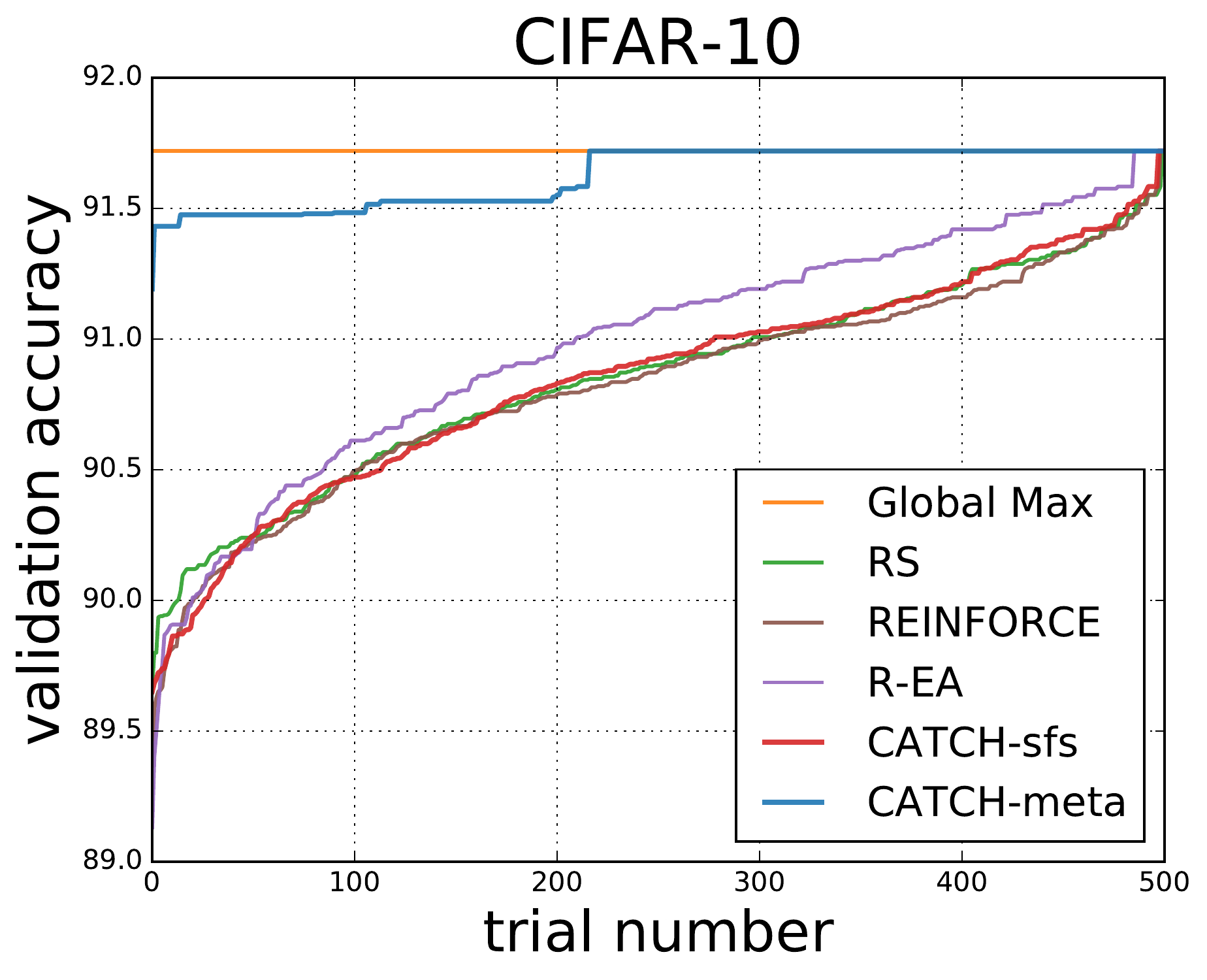}

}\subfloat[CIFAR-100]{\includegraphics[width=0.33\columnwidth]{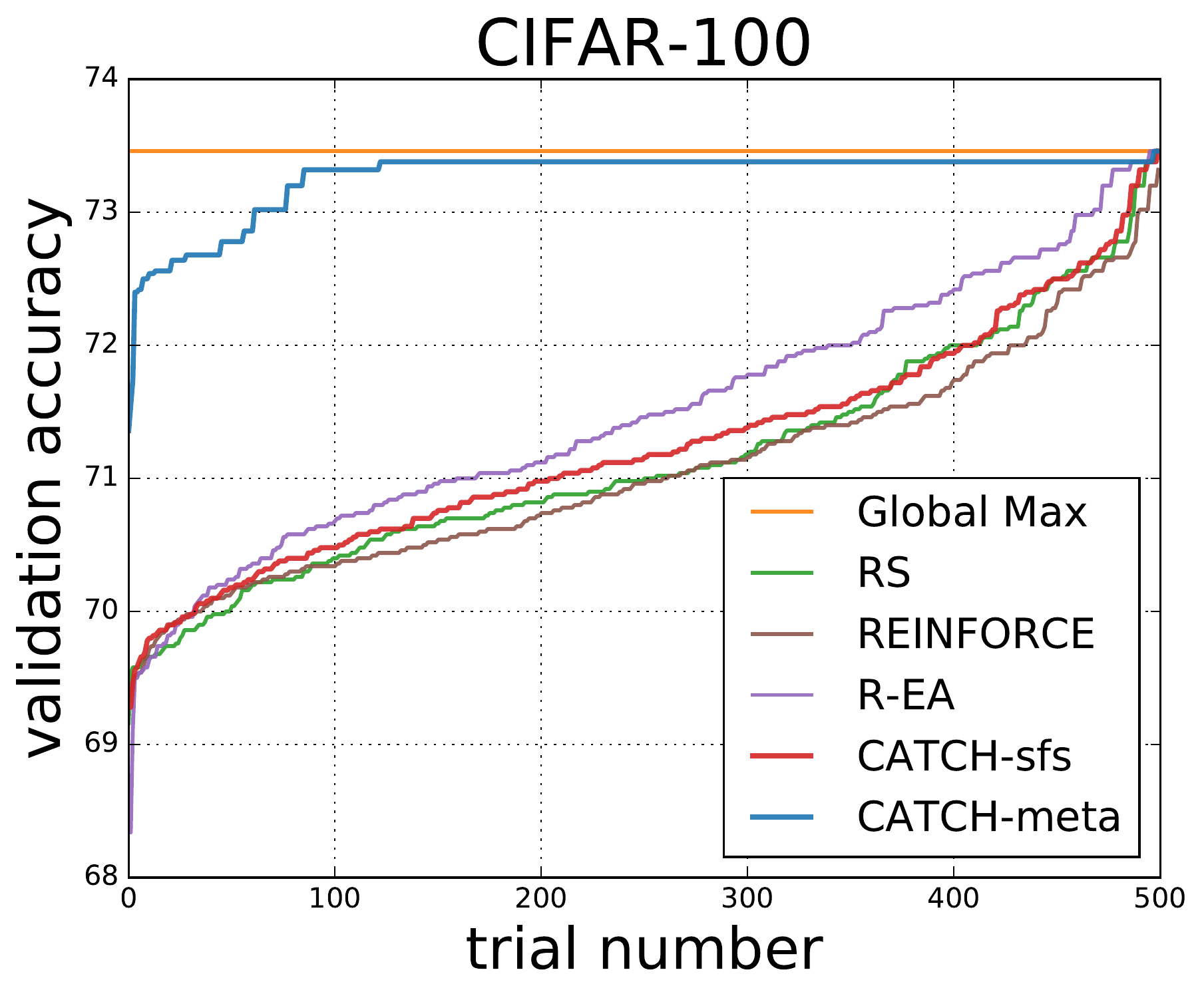}

}\subfloat[ImageNet16-120]{\includegraphics[width=0.33\columnwidth]{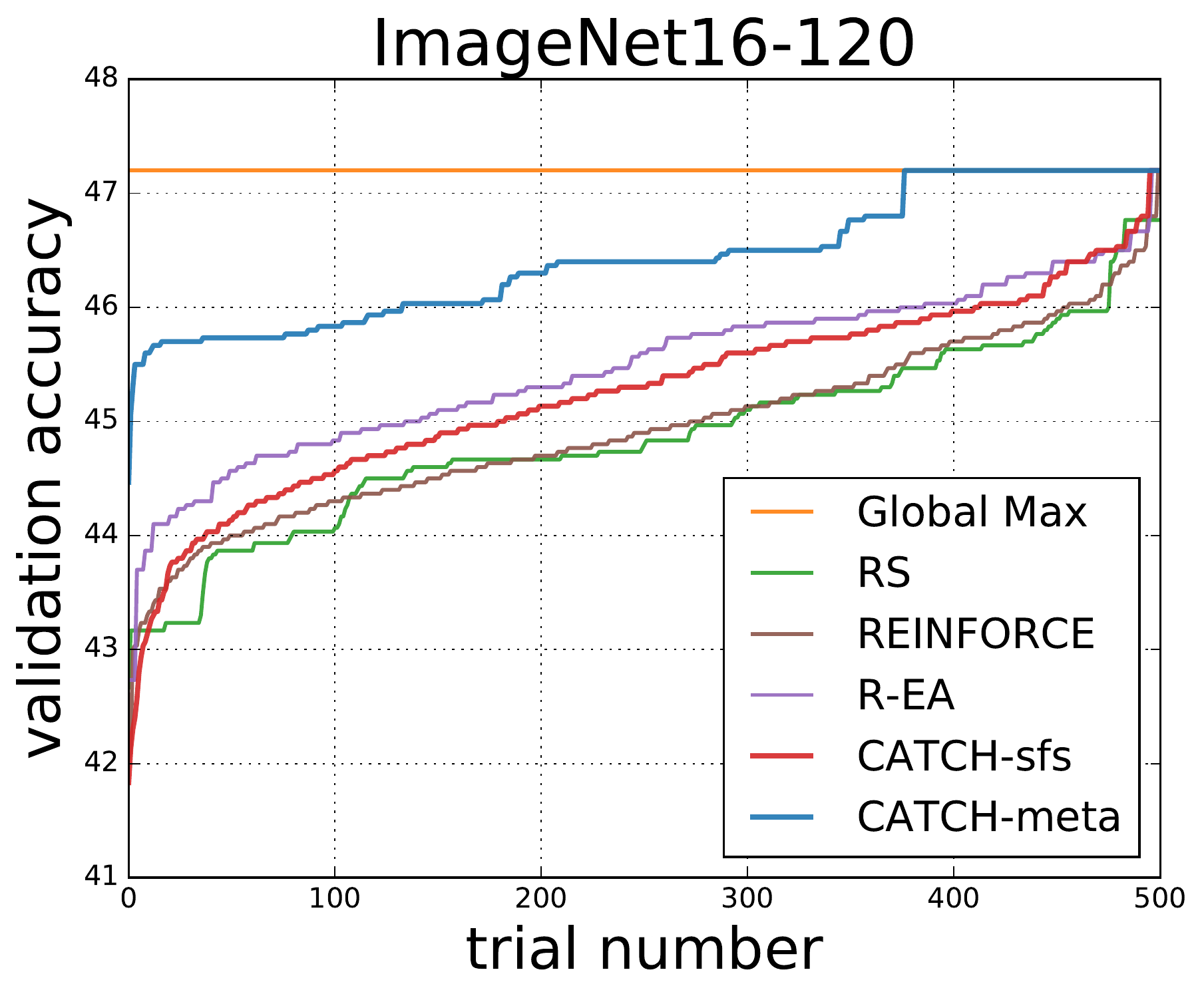}

}
\par\end{centering}

\caption{(a)-(c) show the results of 500 trials for CATCH-meta, CATCH-sfs(search
from scratch) and other sample-based algorithms. Each individual trial
is sorted by the final validation accuracy of the searched network.}

\label{t_graphs}

\end{figure}

\subsection{Implementation Details}

We use Multi-layer Perceptrons (MLP) as the controller policy network to generate
the probability of choosing a certain action. The parameters $\theta_{c}$
of the controller is trained on-policy via the PPO algorithm. We mask invalid actions by zeroing out their probabilities in the  controller\textquoteright s outputs, then softmax the remaining probabilities and sample actions accordingly.

The evaluator is an MLP to generate the predicted score of a network.
In the meta-training phase, we reset $\epsilon$ in the $\epsilon$-greedy exploration strategy each time when the agent initializes a new task. We sample 80\% of
the entries as a batch from the replay buffer using PER.

The encoder MLP outputs a 10-dim latent context vector $\boldsymbol{z}$, and the weight of the KL-Divergence $\beta$ in the combined loss is set to be 0.1. More details of the components' hyperparameters can be found in the Appendix.

\subsection{Benchmark on NAS-Bench-201}

\begin{figure}[t]
\begin{centering}

\par\end{centering}
\begin{centering}
\subfloat[CIFAR-10]{\includegraphics[height=0.26\columnwidth]{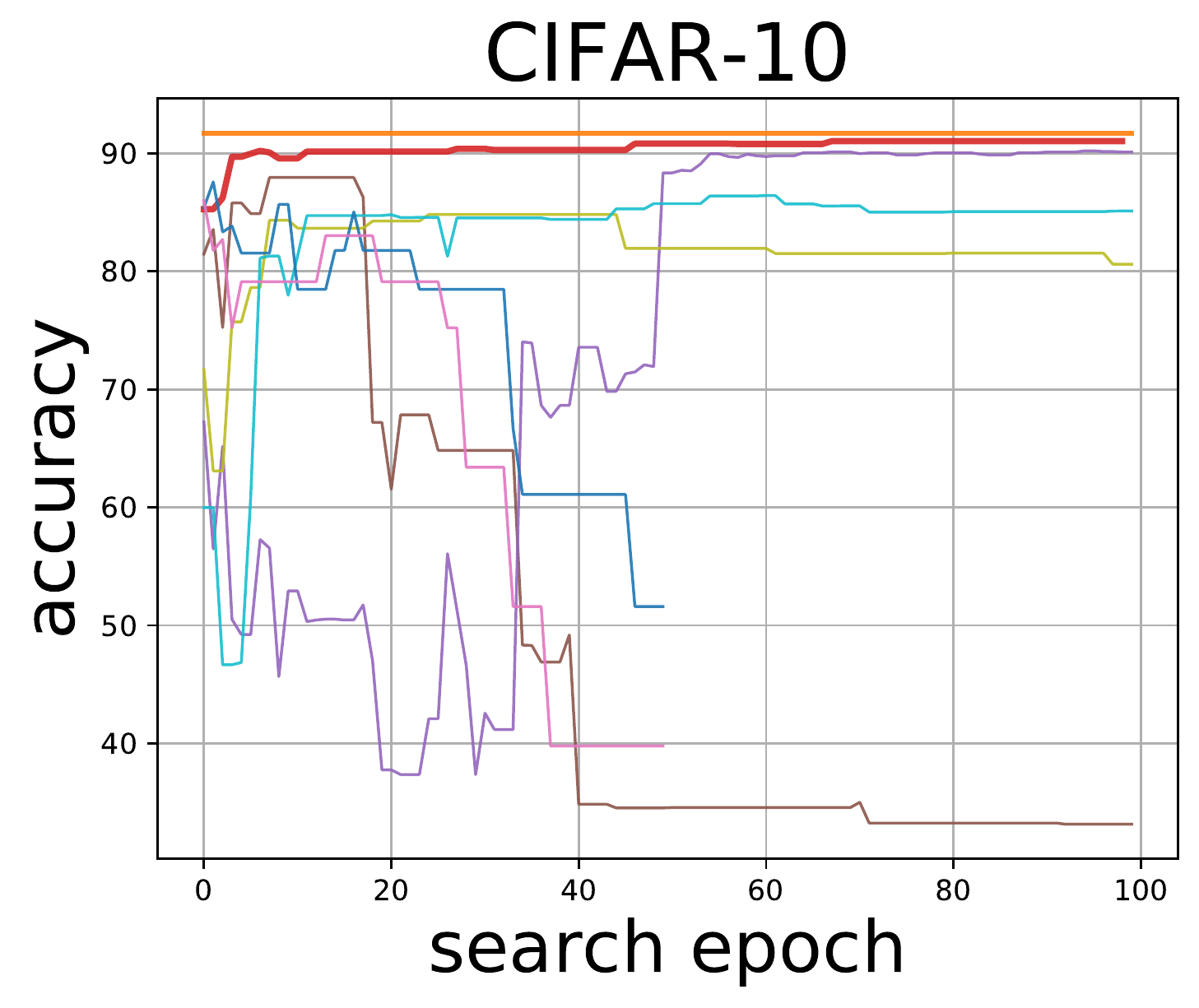}

}\subfloat[CIFAR-100]{\includegraphics[height=0.26\columnwidth]{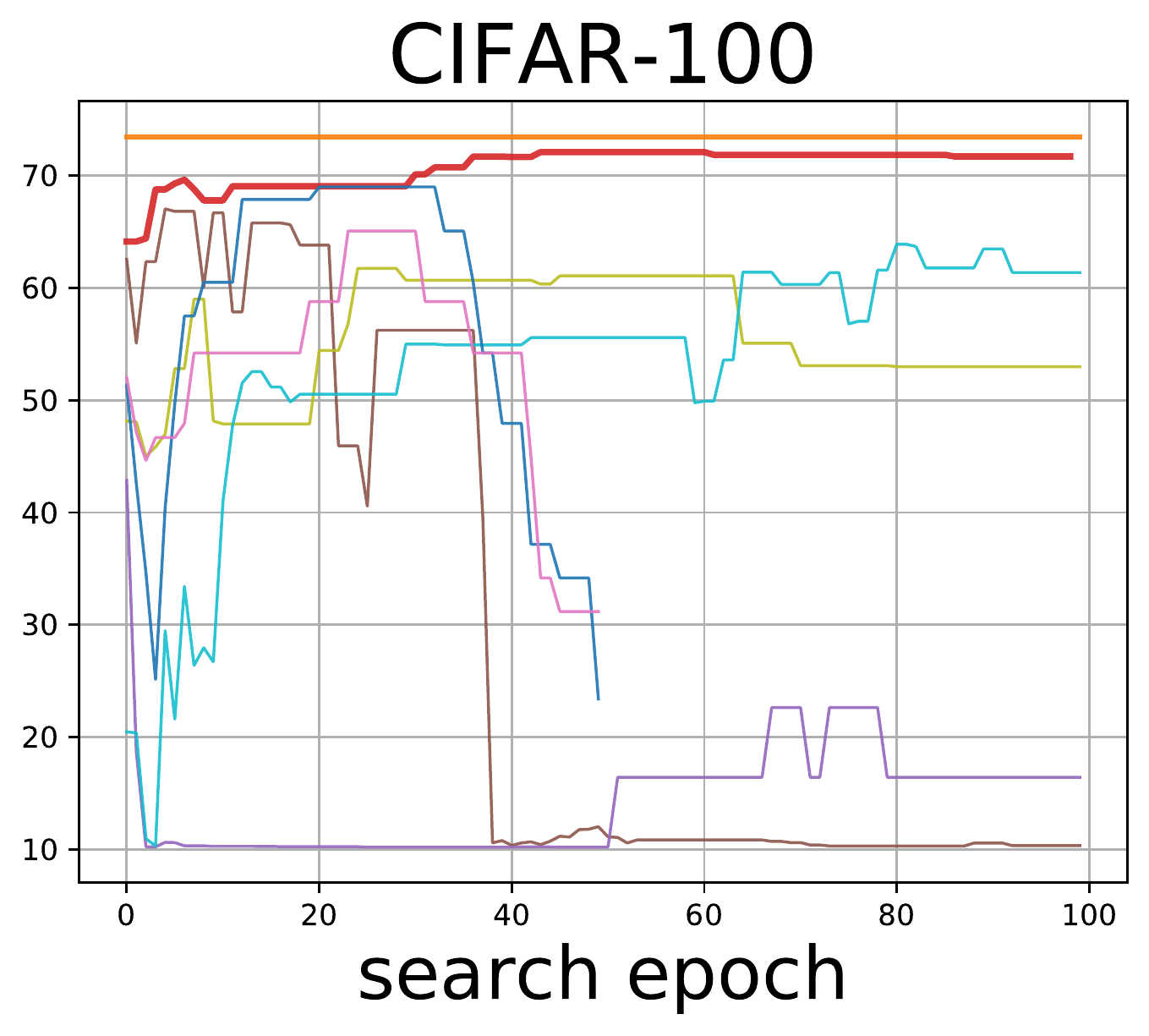}

}\subfloat[ImageNet16-120]{\includegraphics[height=0.26\columnwidth]{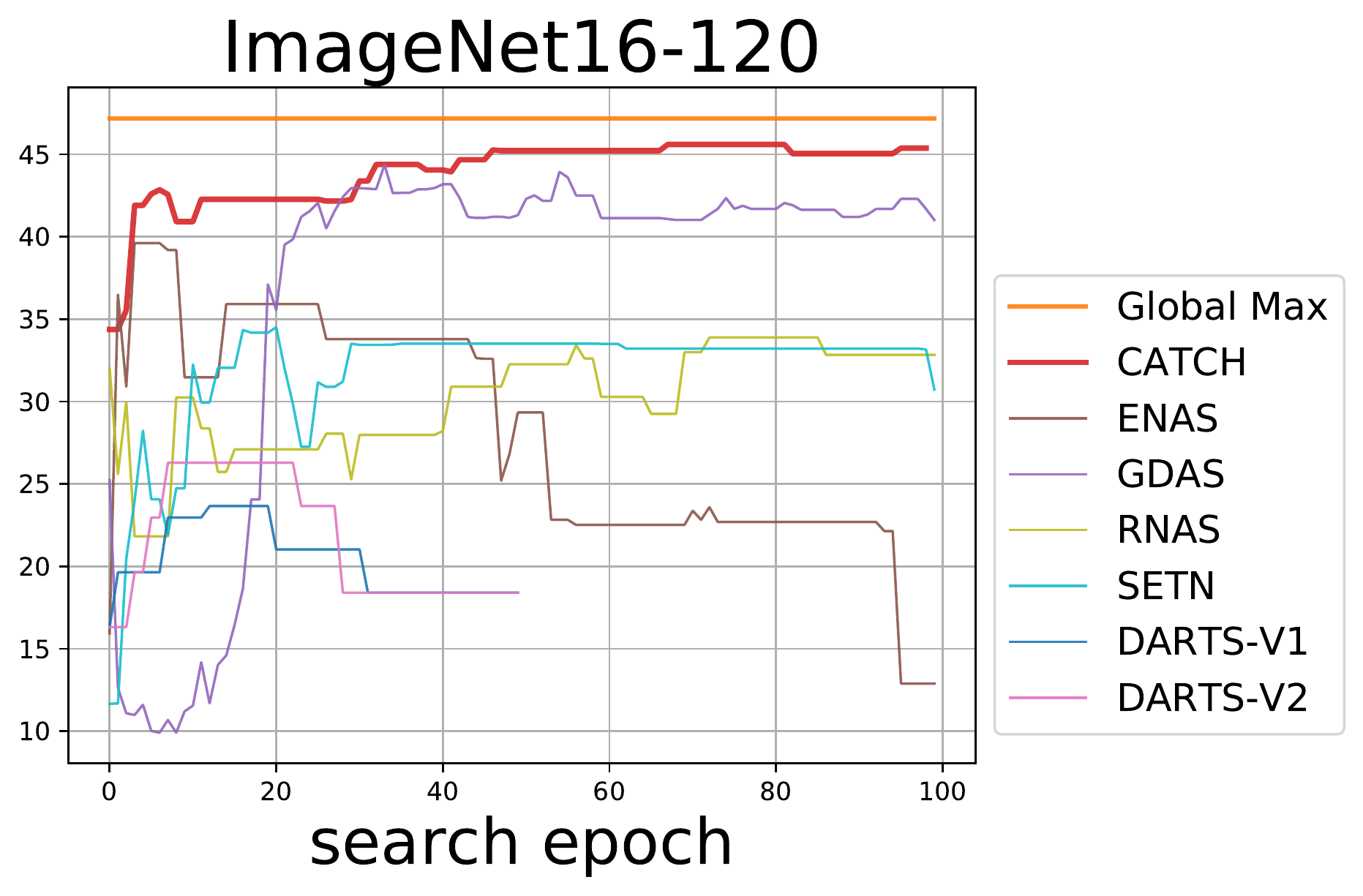}

}
\par\end{centering}
\caption{Learning curves of one-shot algorithms and CATCH. Each curve is an
average of three runs. We plot the first 100 search epochs for algorithms
except for DARTS, which is trained only for 50 search epochs.}
\label{one-shot-lc}
\end{figure}

As recent work \cite{Yang2020NAS} indicated, NAS algorithms are usually
compared unfairly under different settings. To mitigate such problems,
we first tested CATCH on NAS-Bench-201. It is a benchmark dataset
that enables fair comparisons among NAS methods under the same configurations. It supports searching
over cell-based architectures, where a directed acyclic graph represents
each cell with 4 nodes and 5 possible connection operations on each
edge. It provides the validation and test accuracies of 15,625 architectures
on CIFAR-10, CIFAR-100, and ImageNet16-120 datasets. ImageNet16-120
is a subdataset for ImageNet, which downsampled all its images to
$16\times16$, and contains only the first 120 classes of ImageNet.

\subsubsection{Experiment Settings.}

In the meta-training phase, each task is formed as a classification task on an $X$-class sub-dataset of ImageNet16 (ImageNet downsampled to $16\times16$) to maintain consistency with the configurations in NAS-Bench-201. The number of classes $X$ $\in$ $[10,20,30]$. In each meta-epoch, the agent searches 20
networks whose validation accuracies after 12 training epochs are used
as the reward signals. The hyperparameters used for training the networks
in both phases are identical to those in NAS-Bench-201. In the following
experiments, CATCH-meta is meta-trained with 25 meta epochs for 10.5
GPU hours on Tesla V100. We apply
the same configurations as those in NAS-Bench-201.

\subsubsection{Comparison with Sample-based Algorithms.}

We display the search results of the meta-trained version (CATCH-meta)
and the search-from-scratch version (CATCH-sfs where the meta-training
phase is skipped) of our method, and compare them with other sample-based
algorithms: Random Search (RS) \cite{bergstra2012random}, Regularized
Evolution Algorithm (R-EA) \cite{real2019regularized}, and REINFORCE \cite{williams1992simple}.
The results of other methods are reproduced by running the code and
configurations originally provided by NAS-bench-201. Each experiment
is repeated for 500 trials with different seeds. The algorithms are trained for 50 search
epochs in each trial. Figure \ref{t_graphs} presents the search results
on CIFAR-10, CIFAR-100, ImageNet16-120, with the highest validation accuracy on each task.

The reproduced results are consistent with the experiments performed
in NAS-Bench-201. The performance of CATCH-sfs is similar to the other
four methods, but CATCH-meta dominates all other algorithms in the
searched network accuracies. On CIFAR-10, CATCH-meta finds the best
model in 280/500 trials. On CIFAR-100, over half of them find top-3
performance networks within 50 samples, while other algorithms barely
touch the roof. On ImageNet16-120, CATCH reaches the best network
for more than 22\% trials. We
can see tremendous benefits for using the meta-trained CATCH to reduce
time and cost.

\begin{table}[t]
\caption{Comparison of CATCH with one-shot algorithms. The top accuracies of
identified models, standard deviations, search time (hour), total
search time (hour), and the highest validation accuracies among all
the networks in NAS-Bench-201 are reported. The same three random
seeds are used to run through each algorithm. The time budget for
search on CIFAR-10, CIFAR-100, and ImageNet16-120 are 3, 4, and 5
hours respectively.\label{tab:Comparison-one-shot}}

\begin{centering}
\begin{tabular}{c|c|c|c|c|c|c|c}
\hline
\multirow{1}{*}{Algorithm} & \multicolumn{2}{c|}{CIFAR-10} & \multicolumn{2}{c|}{CIFAR-100} & \multicolumn{2}{c|}{ImageNet16-120} & \multirow{2}{*}{Total Time}\tabularnewline
\cline{1-7} \cline{2-7} \cline{3-7} \cline{4-7} \cline{5-7} \cline{6-7} \cline{7-7}
 & Acc $\pm$std & Time & Acc$\pm$std & Time & Acc$\pm$std & Time & \tabularnewline
\hline
DARTS-V1 \cite{liu2018darts} & 88.08$\pm $1.89 & 2.46 & 68.99$\pm$1.93 & 2.44 & 23.66$\pm$0 & 4.55 & 9.45\tabularnewline
DARTS-V2 \cite{liu2018darts} & 87.16$\pm $0.39 & 9 & 65.06$\pm$2.95 & 7.91 & 26.29$\pm$0 & 22.14 & 39.05\tabularnewline
GDAS \cite{dong2019searching} & 90.32$\pm $0.08 & 6 & 70.33$\pm$0.85 & 6.23 & 44.81$\pm$0.97 & 17 & 29.23\tabularnewline
R-NAS \cite{li2019random} & 90.45$\pm $0.43 & 2.19 & 70.39$\pm$1.36 & 2.26 & 44.12$\pm$1.04 & 5.94 & 10.39\tabularnewline
ENAS \cite{pham2018efficient} & 90.2$\pm $0.63 & 4.22 & 69.99$\pm$1.03 & 4.26 & 44.92$\pm$0.51 & 5.18 & 13.66\tabularnewline
SETN \cite{dong2019one} & 90.26$\pm $0.75 & 7.62 & 68.01$\pm$0.21 & 7.74 & 41.04$\pm$1.64 & 20.33 & 35.69\tabularnewline
CATCH-meta & \textbf{91.33$\pm $0.07} & \textbf{3} & \textbf{72.57$\pm$0.81} & \textbf{4} & \textbf{46.07$\pm$0.6} & \textbf{5} & 22.5\tabularnewline
\hline
Max Acc. & \multicolumn{2}{c|}{91.719} & \multicolumn{2}{c|}{73.45} & \multicolumn{2}{c|}{47.19} & ---\tabularnewline
\hline
\end{tabular}
\par\end{centering}
\end{table}

\subsubsection{Comparison with One-shot Algorithms.}

One of the central controversies around meta-NAS algorithms is: given
the high searching efficiency of one-shot methods, can sample-based
algorithms outperform them? We therefore compare the performance of
CATCH with many state-of-the-art one-shot NAS solutions. For fair
comparisons, instead of querying the NAS-Bench-201 network database,
we train each child network for 12 epochs and obtain their early-stop
validation accuracies as training feedbacks. The early-stop training
setup is the same as the one in the meta-training phase. The one-shot
algorithms involved are first-order DARTS (DARTS-V1) \cite{liu2018darts},
second-order DARTS (DARTS-V2), GDAS \cite{dong2019searching}, Random
NAS (R-NAS) \cite{li2019random}, ENAS \cite{pham2018efficient}, and
SETN \cite{dong2019one}. We run the algorithms with the original code
and configurations released from NAS-Bench-201. DARTS-V1 and DARTS-V2
are run for 50 search epochs, and other algorithms are trained for
250 search epochs.

Figure \ref{one-shot-lc} presents the learning curves of each algorithm
in the first 100 search epochs. For CATCH, at each search epoch, we
identify networks with the best partially trained accuracy found so
far, and report their fully trained accuracies. Both DARTS and ENAS
have a relatively strong performance at the beginning, but the curves
drop significantly afterward. SETN resembles Random NAS a lot. GDAS
is among the best one-shot algorithms, but it seems to plateau at
local maximums after a few search epochs. CATCH has the best performance
among all, as it quickly adapts and identifies promising architectures
that are beyond other algorithms' search capacity.

In Table \ref{tab:Comparison-one-shot}, we report the best fully trained
accuracy of networks that each algorithm identifies over their complete
training process. We set the time budget for CATCH to search on CIFAR-10,
CIFAR-100, and ImageNet16-120 as 3, 4, and 5 hours. It is roughly
equivalent to cutting the search on these tasks at 70, 50, and
40 search epochs, respectively. Although DARTS-V1, R-NAS, and ENAS
spend less time in total, they are highly unstable and the performance
of DARTS and ENAS tends to deteriorate over time. CATCH spends 22.5
(10.5 meta + 12 adaptation) hours on all three tasks, and its searched
networks surpass all other algorithms. The presented results have
proved that CATCH is swiftly adaptive, and it is able to identify
networks beyond many one-shot algorithms\textquoteright{} reach within
a reasonable time.

\subsection{Experiments on Residual Block-based Search Space}

\begin{table}[t]
\caption{Results on ImageNet compared to manually designed and NAS searched
architectures. Latency is measured on one Tesla V100 with one image
with shape (3, 720, 1080).}

\begin{centering}
\begin{tabular}{c|c|c|c}
\hline
Network & Top-1 Acc (\%) & Top-5 Acc (\%) & Latency (ms)\tabularnewline
\hline
ResNet50 \cite{he2016deep} & 77.15 & 93.29 & 16.4\tabularnewline
DenseNet201 \cite{huang2017densely} & 77.42 & 93.66 & 31.6\tabularnewline
ResNext101 \cite{xie2017aggregated} & 79.31 & 94.5 & 76.7\tabularnewline
Inception-V3 \cite{szegedy2016rethinking} & 78.8 & 94.4 & 16.4\tabularnewline
\hline
EfficientNet-B1 \cite{tan2019efficientnet} & 77.3 & 93.5 & 29.5\tabularnewline
EfficientNet-B2 & 79.2 & 94.5 & 47.6\tabularnewline
NASNet-A \cite{zoph2018learning} & 78.6 & 94.2 & -\tabularnewline
BASE \cite{shaw2019meta} & 74.3 & 91.9 & -\tabularnewline
\hline
CATCH-Net-A & 79.04 & 94.43 & \textbf{16.9}\tabularnewline
CATCH-Net-B & \textbf{79.46} & \textbf{94.7} & 33.7\tabularnewline
\hline
\end{tabular}
\par\end{centering}
\label{imagenet table}
\end{table}

Having proved that CATCH can adapt to new tasks efficiently with
meta-training, we further inquire whether CATCH has the ability to
transfer across different domains including image classification, objection
detection, and semantic segmentation. In this section, we consider
a more challenging setting where the meta-training phase contain
only image classification tasks while tasks in all the three domains are targeted
in the adaptation phase. The architectures are very different among
these domains, so we search for their common component - the feature
extractor (backbone).
ResNet is one popular backbone for these tasks, thus we design the search space following \cite{DBLP:journals/corr/abs-1906-04423,yao2019sm}.

Constructing a model in the Residual block-based search space requires
the controller to make several decisions: (1) select the network's
base channel from $[48,56,64,72]$, (2) decide the network's depth
within $[15,20,25,30]$, (3) choose the number of stages $s$, which
is either 4 or 5, (4) schedule the number of blocks contained in each
stage, and (5) arrange the distribution of blocks holding different
channels. Details of the Residual block-based search space can be
found in the Appendix.

\begin{table}[t]
\caption{Results on COCO compared to manually designed and NAS searched backbones.
Latency results of networks except CATCH are referred from \cite{yao2019sm}.}

\begin{centering}
\begin{tabular}{c|c|c|c|c}
\hline
Method & Backbone & Input size & Latency (ms) & mAP\tabularnewline
\hline
RetinaNet \cite{lin2017focal} & ResNet101-FPN & 1333x800 & 91.7 (V100) & 39.1\tabularnewline
FSAF \cite{zhu2019feature} & ResNet101-FPN & 1333x800 & 92.5 (V100) & 40.9\tabularnewline
GA-Faster RCNN \cite{wang2019region} & ResNet50-FPN & 1333x800 & 104.2 (V100) & 39.8\tabularnewline
Faster-RCNN \cite{ren2015faster} & ResNet101-FPN & 1333x800 & 84.0 (V100) & 39.4\tabularnewline
Mask-RCNN \cite{he2017mask} & ResNet101-FPN & 1333x800 & 105.0 (V100) & 40.2\tabularnewline
\hline
DetNAS \cite{chen2019detnas} & Searched Backbone & 1333x800 & - & 42.0\tabularnewline
SM-NAS: E3 & Searched Backbone & 800x600 & 50.7(V100) & 42.8\tabularnewline
SM-NAS: E5 & Searched Backbone & 1333x800 & 108.1(V100) & 45.9\tabularnewline
Auto-FPN \cite{xu2019auto} & Searched Backbone & 1333x800 & - & 40.5\tabularnewline
\hline
CATCH & CATCH-Net-C & 1333x800 & 123.5 (V100) & \textbf{43.2}\tabularnewline
\hline
\end{tabular}
\par\end{centering}
\label{coco}

\end{table}

\subsubsection{Experiment Settings.}

We use the same meta-training settings as the ones we used in NAS-Bench-201.
For each meta epoch, an ImageNet sub-dataset is created. To form such sub-datasets, we sample $X$ classes from all classes of ImageNet, where $X\in [10,20,30]$. Then the images are resize to $16\times16$,
$32\times32$, or $224\times224$. Thus there are $3\times\left[\tbinom{1000}{10}+\tbinom{1000}{20}+\tbinom{1000}{30}\right]$
possible sub-datasets.

To achieve the balance between inference latency and network performance,
we adopt the multi-objectve reward function $R=P(m)\times[\frac{LAT(m)}{T_{target}}]^{w}$
in \cite{Tan_2019_CVPR}, where $P(m)$ denotes the model \textquoteright s performance
(e.g. validation accuracy for classification, mAP for object detection or mIoU for semantic segmentation), $LAT(m)$ measures the model's inference latency, and $T_{target}$ is the target latency. $w$ serves as a hyperparameter
adjusting the performance-latency tradeoff.
In our experiments, we set $w=-0.05$. With this reward, we hope to
find models that excel not only in performance but also in inference
speed. We meta train CATCHer for 5 GPU days, and adapt on each target task
to search for 10 architectures. We target ImageNet dataset for image classification,
COCO dataset for object detection and Cityscapes dataset for semantic segmentation.
The detailed settings can be found in the Appendix.

\subsubsection{Search Results.}

\begin{table}[t]
\begin{centering}
\caption{Results on Cityscapes compared to manually designed and NAS searched
backbones. Latency is measured on Tesla V100 with one image with shape
(3, 1024, 1024). SS and MS denote for single scale and multiple scale
testing respectively.\label{Tab:cityscapes_val}}
\par\end{centering}
\begin{centering}
\begin{tabular}{c|c|c|c|c}
\hline
Method & Backbone & Latency (ms) & mIoU (SS) & mIoU (MS)\tabularnewline
\hline
BiSeNet \cite{Yu_2018_ECCV} & ResNet101 & 41 & - & 80.3\tabularnewline
DeepLabv3+ \cite{Chen_2018_ECCV} & Xception-65 & 85 & 77.82 & 79.3\tabularnewline
CCNet \cite{Huang_2019_ICCV} & ResNet50 & 175 & - & 78.5\tabularnewline
DUC \cite{8354267} & ResNet152 & - & 76.7 & -\tabularnewline
DANet \cite{Fu_2019_CVPR} & ResNet50 & - & 76.34 & -\tabularnewline
\hline
Auto-DeepLab \cite{liu2019auto} & Searched Backbone & - & 79.94 & -\tabularnewline
DPC \cite{NIPS2018_8087} & Xception-71 & - & 80.1 & -\tabularnewline
\hline
CATCH & CATCH-Net-D & \textbf{27} & 79.52 & \textbf{81.12}\tabularnewline
\hline
\end{tabular}
\par\end{centering}
\end{table}

Table \ref{imagenet table} compares the searched architectures with
other widely-recognized networks on ImageNet. CATCH-Net-A outperforms
many listed networks. Its accuracy is comparable with EfficientNet-B1
and ResNext-101, yet it is 2.82X and 4.54X faster. CATCH-Net-B
outperforms ResNext-101 while shortens the latency by 2.28X. The network
comparison on COCO and Cityscapes is presented in Table \ref{coco}
and Table \ref{Tab:cityscapes_val}. Our network again shows faster
inference time and competitive performance. We also transfer CATCH-Net-B
found during the search on ImageNet to COCO and Cityscapes, which
yield 42\% mAP with 136ms inference time and 80.87\% mIoU (MS) with
52ms latency, respectively. Our results again show that directly transferring
top architectures from one task to another cannot guarantee optimality.
It also reveals CATCH\textquoteright s potentials to transfer across
tasks even when they are distant from the meta-training ones.

\section{Ablation Study}
The context encoder is the spotlight component of our algorithm. We
are especially curious about: (1) Is the encoder actually helpful
for adaptation (compared with simply plugging in the meta-learned
controller and evaluator priors)? (2) If so, does the improvement
come from good estimates of the posterior, or is it from the stochastic
generation of $\boldsymbol{z}$ that encourages exploration and benefits
generalization?

\begin{figure}[t]

\begin{centering}
\subfloat[]{\includegraphics[scale=0.21]{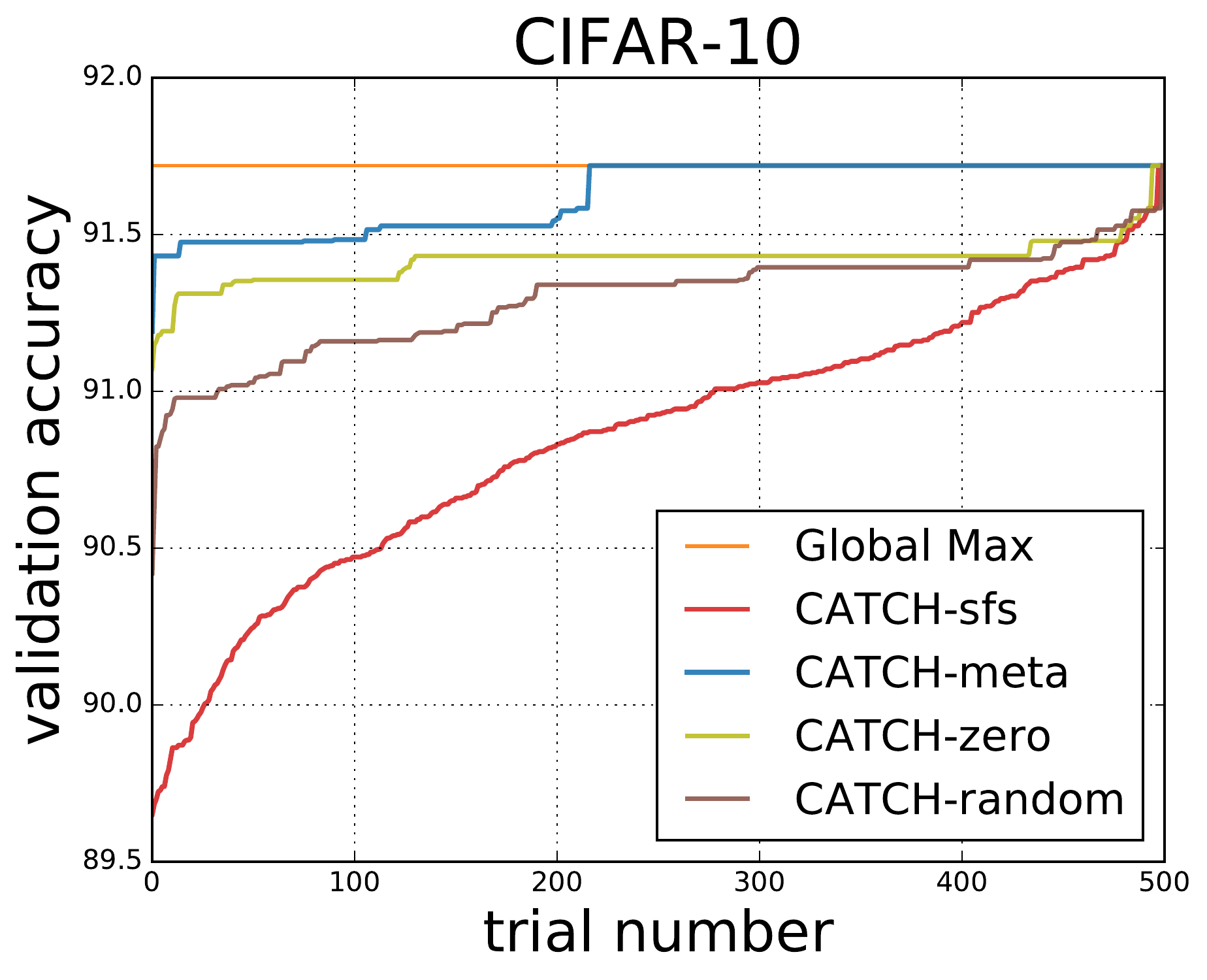}

}\subfloat[]{\includegraphics[scale=0.21]{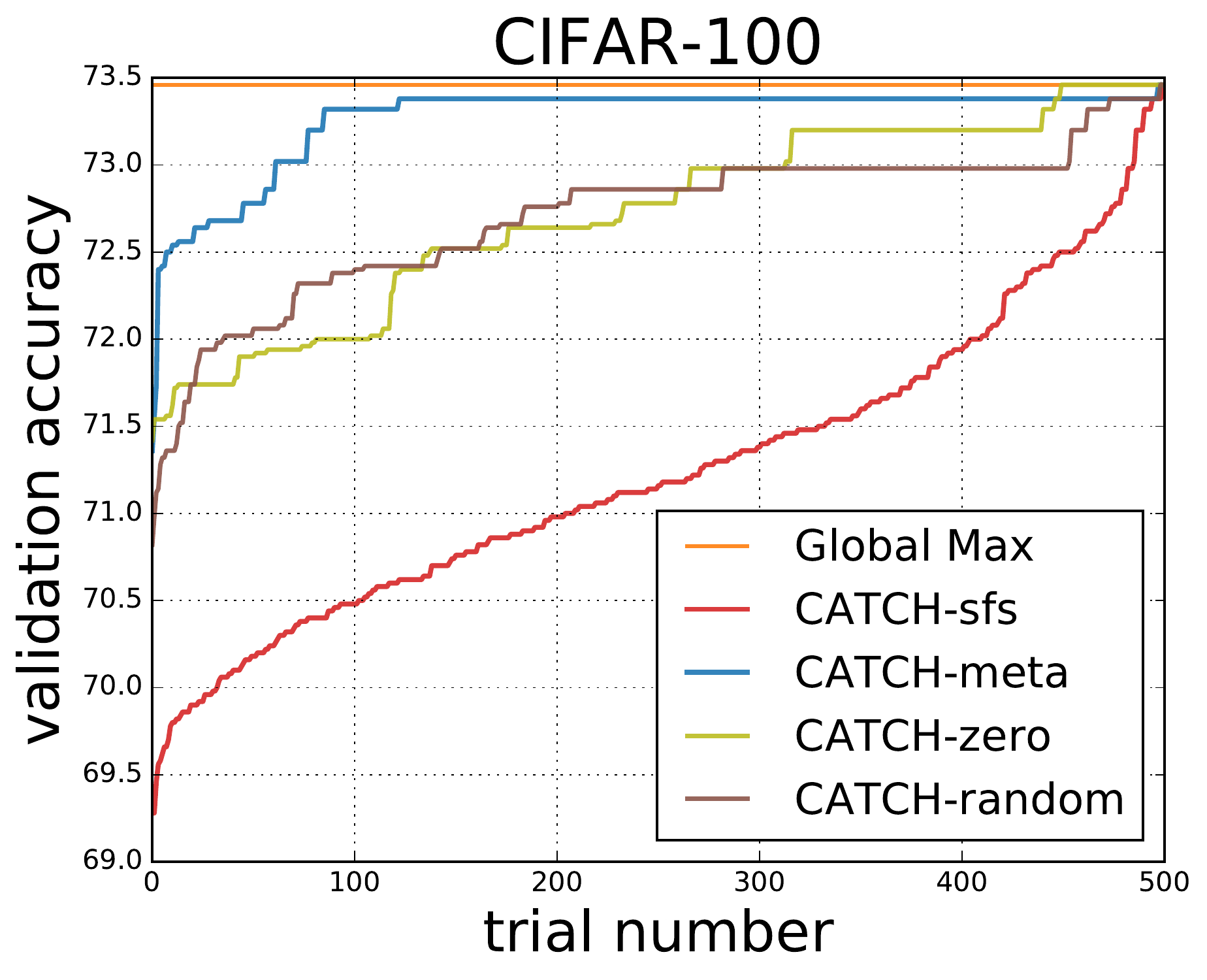}

}\subfloat[]{\includegraphics[scale=0.21]{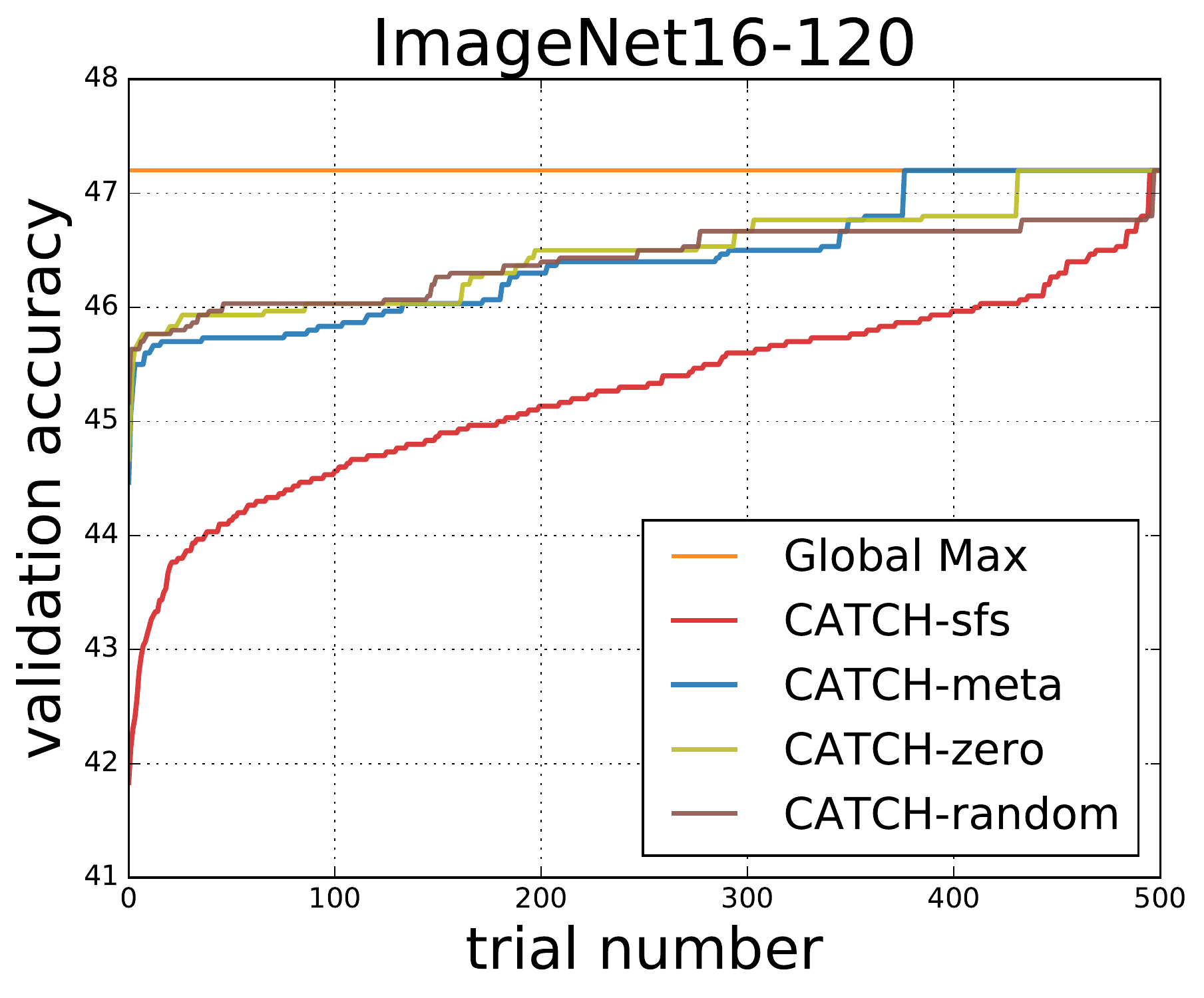}

}
\par\end{centering}

\caption{(a)-(c) compare results of 500 trials for CATCH-meta, CATCH-sfs(search
from scratch), CATCH-zero, CATCH-random.}

\label{t_graphs-ab-enc}

\end{figure}

To answer these questions, we designed two extra sets of experiments:
(1) CATCH-zero: We set $\boldsymbol{z}=\boldsymbol{0}$, and thereby
completely eliminate the encoder's effect on both the controller and
the evaluator; (2) CATCH-random: We sample each $\boldsymbol{z}$ from a unit
Gaussian prior $\mathcal{N}(\boldsymbol{0},diag(\boldsymbol{1}))$ during the search as random
inputs. The results are presented in Figure \ref{t_graphs-ab-enc}
(a)-(c). In both settings, the agents are still meta-trained for 10.5
hours before they are plugged in for adaptation.

The gaps among the lines in Figure \ref{t_graphs-ab-enc}
answered our questions. The encoder not only helps with adaptation
(through comparing CATCH-meta and CATCH-zero), but also provides assistance
in a much more meaningful way than using random inputs for exploration,
as CATCH-meta outperforms CATCH-random on both CIFAR-10 and CIFAR-100. Interestingly,
we observe less significant improvement on ImageNet16-120. One hypothesis is since we perform the meta-training phase on sub-datasets
of ImageNet16, the meta-trained controller and evaluator are already
tuned towards policies that fit the search on ImageNet16. Hence, the
transferred policies require less adaptation assistance from the encoder. More ablation studies can be found in the Appendix.

\section{Conclusion and Discussion}

In this work, we propose CATCH, a transferrable NAS approach, by designing
an efficient learning framework that leverages the benefits of context-based
meta reinforcement learning. The key contribution of CATCH is to boost
NAS efficiency by extracting and utilizing task-specific latent contexts,
while maintaining universality and robustness in various settings.
Experiments and ablation studies show its dominant position in search
efficiency and performance over non-transferrable schemes on NAS-Bench-201.
Extensive experiments on residual block-based search space also demonstrate
its capability in handling cross-task architecture search. As a task-agnostic
transferrable NAS framework, CATCH possesses great potentials in scaling
 NAS to large datasets and various domains efficiently.

During our research into transferrable NAS frameworks, we identified
many potentially valuable questions to be explored. Efficient adaptation
among domains is challenging, and we demonstrated a first attempt to
simplify it by searching for backbones with a shared search space.
A possible future investigation would be to generalize cross-task
architecture search to flexibly include more decisions, such as searching
for detection and segmentation heads. Meanwhile, our meta-training
tasks involve only classification tasks, but it is also possible
to diversify the pool and explore whether it leads to further performance
boosts.

\bibliographystyle{plain}
\bibliography{3177}

\begin{thebibliography}{10}

\bibitem{alemi2016deep}
Alexander~A Alemi, Ian Fischer, Joshua~V Dillon, and Kevin Murphy.
\newblock Deep variational information bottleneck.
\newblock {\em arXiv preprint arXiv:1612.00410}, 2016.

\bibitem{bergstra2012random}
James Bergstra and Yoshua Bengio.
\newblock Random search for hyper-parameter optimization.
\newblock {\em Journal of machine learning research}, 13(Feb):281--305, 2012.

\bibitem{bergstra2013making}
James Bergstra, Daniel Yamins, and David Cox.
\newblock Making a science of model search: Hyperparameter optimization in
  hundreds of dimensions for vision architectures.
\newblock In {\em International conference on machine learning}, pages
  115--123, 2013.

\bibitem{bergstra2011algorithms}
James~S Bergstra, R{\'e}mi Bardenet, Yoshua Bengio, and Bal{\'a}zs K{\'e}gl.
\newblock Algorithms for hyper-parameter optimization.
\newblock In {\em Advances in neural information processing systems}, pages
  2546--2554, 2011.

\bibitem{chen2018searching}
Liang-Chieh Chen, Maxwell Collins, Yukun Zhu, George Papandreou, Barret Zoph,
  Florian Schroff, Hartwig Adam, and Jon Shlens.
\newblock Searching for efficient multi-scale architectures for dense image
  prediction.
\newblock In {\em Advances in Neural Information Processing Systems}, pages
  8699--8710, 2018.

\bibitem{NIPS2018_8087}
Liang-Chieh Chen, Maxwell Collins, Yukun Zhu, George Papandreou, Barret Zoph,
  Florian Schroff, Hartwig Adam, and Jon Shlens.
\newblock Searching for efficient multi-scale architectures for dense image
  prediction.
\newblock In S.~Bengio, H.~Wallach, H.~Larochelle, K.~Grauman, N.~Cesa-Bianchi,
  and R.~Garnett, editors, {\em Advances in Neural Information Processing
  Systems 31}, pages 8699--8710. Curran Associates, Inc., 2018.

\bibitem{Chen_2018_ECCV}
Liang-Chieh Chen, Yukun Zhu, George Papandreou, Florian Schroff, and Hartwig
  Adam.
\newblock Encoder-decoder with atrous separable convolution for semantic image
  segmentation.
\newblock In {\em The European Conference on Computer Vision (ECCV)}, September
  2018.

\bibitem{chen2019detnas}
Yukang Chen, Tong Yang, Xiangyu Zhang, Gaofeng Meng, Chunhong Pan, and Jian
  Sun.
\newblock Detnas: Neural architecture search on object detection.
\newblock {\em arXiv preprint arXiv:1903.10979}, 2019.

\bibitem{Cordts2016Cityscapes}
Marius Cordts, Mohamed Omran, Sebastian Ramos, Timo Rehfeld, Markus Enzweiler,
  Rodrigo Benenson, Uwe Franke, Stefan Roth, and Bernt Schiele.
\newblock The cityscapes dataset for semantic urban scene understanding.
\newblock In {\em Proc. of the IEEE Conference on Computer Vision and Pattern
  Recognition (CVPR)}, 2016.

\bibitem{deng2009imagenet}
Jia Deng, Wei Dong, Richard Socher, Li-Jia Li, Kai Li, and Li~Fei-Fei.
\newblock Imagenet: A large-scale hierarchical image database.
\newblock In {\em 2009 IEEE conference on computer vision and pattern
  recognition}, pages 248--255. Ieee, 2009.

\bibitem{dong2019one}
Xuanyi Dong and Yi~Yang.
\newblock One-shot neural architecture search via self-evaluated template
  network.
\newblock In {\em Proceedings of the IEEE International Conference on Computer
  Vision}, pages 3681--3690, 2019.

\bibitem{dong2019searching}
Xuanyi Dong and Yi~Yang.
\newblock Searching for a robust neural architecture in four gpu hours.
\newblock In {\em Proceedings of the IEEE Conference on Computer Vision and
  Pattern Recognition}, pages 1761--1770, 2019.

\bibitem{dong2020bench}
Xuanyi Dong and Yi~Yang.
\newblock Nas-bench-201: Extending the scope of reproducible neural
  architecture search.
\newblock {\em arXiv preprint arXiv:2001.00326}, 2020.

\bibitem{elsken2018efficient}
Thomas Elsken, Jan~Hendrik Metzen, and Frank Hutter.
\newblock Efficient multi-objective neural architecture search via lamarckian
  evolution.
\newblock {\em arXiv preprint arXiv:1804.09081}, 2018.

\bibitem{elsken2019meta}
Thomas Elsken, Benedikt Staffler, Jan~Hendrik Metzen, and Frank Hutter.
\newblock Meta-learning of neural architectures for few-shot learning.
\newblock {\em arXiv preprint arXiv:1911.11090}, 2019.

\bibitem{finn2017model}
Chelsea Finn, Pieter Abbeel, and Sergey Levine.
\newblock Model-agnostic meta-learning for fast adaptation of deep networks.
\newblock {\em arXiv preprint arXiv:1703.03400}, 2017.

\bibitem{Fu_2019_CVPR}
Jun Fu, Jing Liu, Haijie Tian, Yong Li, Yongjun Bao, Zhiwei Fang, and Hanqing
  Lu.
\newblock Dual attention network for scene segmentation.
\newblock In {\em The IEEE Conference on Computer Vision and Pattern
  Recognition (CVPR)}, June 2019.

\bibitem{he2017mask}
Kaiming He, Georgia Gkioxari, Piotr Doll{\'a}r, and Ross Girshick.
\newblock Mask r-cnn.
\newblock In {\em Proceedings of the IEEE international conference on computer
  vision}, pages 2961--2969, 2017.

\bibitem{he2016deep}
Kaiming He, Xiangyu Zhang, Shaoqing Ren, and Jian Sun.
\newblock Deep residual learning for image recognition.
\newblock In {\em Proceedings of the IEEE conference on computer vision and
  pattern recognition}, pages 770--778, 2016.

\bibitem{huang2017densely}
Gao Huang, Zhuang Liu, Laurens Van Der~Maaten, and Kilian~Q Weinberger.
\newblock Densely connected convolutional networks.
\newblock In {\em Proceedings of the IEEE conference on computer vision and
  pattern recognition}, pages 4700--4708, 2017.

\bibitem{Huang_2019_ICCV}
Zilong Huang, Xinggang Wang, Lichao Huang, Chang Huang, Yunchao Wei, and Wenyu
  Liu.
\newblock Ccnet: Criss-cross attention for semantic segmentation.
\newblock In {\em The IEEE International Conference on Computer Vision (ICCV)},
  October 2019.

\bibitem{huber1992robust}
Peter~J Huber.
\newblock Robust estimation of a location parameter.
\newblock In {\em Breakthroughs in statistics}, pages 492--518. Springer, 1992.

\bibitem{kim2018auto}
Jaehong Kim, Sangyeul Lee, Sungwan Kim, Moonsu Cha, Jung~Kwon Lee, Youngduck
  Choi, Yongseok Choi, Dong-Yeon Cho, and Jiwon Kim.
\newblock Auto-meta: Automated gradient based meta learner search.
\newblock {\em arXiv preprint arXiv:1806.06927}, 2018.

\bibitem{kingma2013auto}
Diederik~P Kingma and Max Welling.
\newblock Auto-encoding variational bayes.
\newblock {\em arXiv preprint arXiv:1312.6114}, 2013.

\bibitem{lan2019meta}
Lin Lan, Zhenguo Li, Xiaohong Guan, and Pinghui Wang.
\newblock Meta reinforcement learning with task embedding and shared policy.
\newblock {\em arXiv preprint arXiv:1905.06527}, 2019.

\bibitem{li2019random}
Liam Li and Ameet Talwalkar.
\newblock Random search and reproducibility for neural architecture search.
\newblock {\em arXiv preprint arXiv:1902.07638}, 2019.

\bibitem{li2017meta}
Zhenguo Li, Fengwei Zhou, Fei Chen, and Hang Li.
\newblock Meta-sgd: Learning to learn quickly for few-shot learning.
\newblock {\em arXiv preprint arXiv:1707.09835}, 2017.

\bibitem{Lian2020Towards}
Dongze Lian, Yin Zheng, Yintao Xu, Yanxiong Lu, Leyu Lin, Peilin Zhao, Junzhou
  Huang, and Shenghua Gao.
\newblock Towards fast adaptation of neural architectures with meta learning.
\newblock In {\em International Conference on Learning Representations}, 2020.

\bibitem{lin2017focal}
Tsung-Yi Lin, Priya Goyal, Ross Girshick, Kaiming He, and Piotr Doll{\'a}r.
\newblock Focal loss for dense object detection.
\newblock In {\em Proceedings of the IEEE international conference on computer
  vision}, pages 2980--2988, 2017.

\bibitem{lin2014microsoft}
Tsung-Yi Lin, Michael Maire, Serge Belongie, James Hays, Pietro Perona, Deva
  Ramanan, Piotr Dollar, and Larry Zitnick.
\newblock Microsoft coco: Common objects in context.
\newblock In {\em ECCV}. European Conference on Computer Vision, September
  2014.

\bibitem{liu2020autofis}
Bin Liu, Chenxu Zhu, Guilin Li, Weinan Zhang, Jincai Lai, Ruiming Tang,
  Xiuqiang He, Zhenguo Li, and Yong Yu.
\newblock Autofis: Automatic feature interaction selection in factorization
  models for click-through rate prediction.
\newblock {\em arXiv preprint arXiv:2003.11235}, 2020.

\bibitem{liu2019auto}
Chenxi Liu, Liang-Chieh Chen, Florian Schroff, Hartwig Adam, Wei Hua, Alan~L
  Yuille, and Li~Fei-Fei.
\newblock Auto-deeplab: Hierarchical neural architecture search for semantic
  image segmentation.
\newblock In {\em Proceedings of the IEEE Conference on Computer Vision and
  Pattern Recognition}, pages 82--92, 2019.

\bibitem{liu2018darts}
Hanxiao Liu, Karen Simonyan, and Yiming Yang.
\newblock Darts: Differentiable architecture search.
\newblock {\em arXiv preprint arXiv:1806.09055}, 2018.

\bibitem{mnih2013playing}
Volodymyr Mnih, Koray Kavukcuoglu, David Silver, Alex Graves, Ioannis
  Antonoglou, Daan Wierstra, and Martin Riedmiller.
\newblock Playing atari with deep reinforcement learning.
\newblock {\em arXiv preprint arXiv:1312.5602}, 2013.

\bibitem{negrinho2017deeparchitect}
Renato Negrinho and Geoff Gordon.
\newblock Deeparchitect: Automatically designing and training deep
  architectures.
\newblock {\em arXiv preprint arXiv:1704.08792}, 2017.

\bibitem{pasunuru2019continual}
Ramakanth Pasunuru and Mohit Bansal.
\newblock Continual and multi-task architecture search.
\newblock {\em arXiv preprint arXiv:1906.05226}, 2019.

\bibitem{pham2018efficient}
Hieu Pham, Melody~Y Guan, Barret Zoph, Quoc~V Le, and Jeff Dean.
\newblock Efficient neural architecture search via parameter sharing.
\newblock {\em arXiv preprint arXiv:1802.03268}, 2018.

\bibitem{rakelly2019efficient}
Kate Rakelly, Aurick Zhou, Deirdre Quillen, Chelsea Finn, and Sergey Levine.
\newblock Efficient off-policy meta-reinforcement learning via probabilistic
  context variables.
\newblock {\em arXiv preprint arXiv:1903.08254}, 2019.

\bibitem{real2019aging}
E~Real, A~Aggarwal, Y~Huang, and QV~Le.
\newblock Aging evolution for image classifier architecture search.
\newblock In {\em AAAI Conference on Artificial Intelligence}, 2019.

\bibitem{real2019regularized}
Esteban Real, Alok Aggarwal, Yanping Huang, and Quoc~V Le.
\newblock Regularized evolution for image classifier architecture search.
\newblock In {\em Proceedings of the AAAI Conference on Artificial
  Intelligence}, volume~33, pages 4780--4789, 2019.

\bibitem{ren2015faster}
Shaoqing Ren, Kaiming He, Ross Girshick, and Jian Sun.
\newblock Faster r-cnn: Towards real-time object detection with region proposal
  networks.
\newblock In {\em Advances in neural information processing systems}, pages
  91--99, 2015.

\bibitem{schaul2015prioritized}
Tom Schaul, John Quan, Ioannis Antonoglou, and David Silver.
\newblock Prioritized experience replay.
\newblock {\em arXiv preprint arXiv:1511.05952}, 2015.

\bibitem{schulman2017proximal}
John Schulman, Filip Wolski, Prafulla Dhariwal, Alec Radford, and Oleg Klimov.
\newblock Proximal policy optimization algorithms.
\newblock {\em arXiv preprint arXiv:1707.06347}, 2017.

\bibitem{shaw2019meta}
Albert Shaw, Wei Wei, Weiyang Liu, Le~Song, and Bo~Dai.
\newblock Meta architecture search.
\newblock In {\em Advances in Neural Information Processing Systems}, pages
  11225--11235, 2019.

\bibitem{szegedy2016rethinking}
Christian Szegedy, Vincent Vanhoucke, Sergey Ioffe, Jon Shlens, and Zbigniew
  Wojna.
\newblock Rethinking the inception architecture for computer vision.
\newblock In {\em Proceedings of the IEEE conference on computer vision and
  pattern recognition}, pages 2818--2826, 2016.

\bibitem{Tan_2019_CVPR}
Mingxing Tan, Bo~Chen, Ruoming Pang, Vijay Vasudevan, Mark Sandler, Andrew
  Howard, and Quoc~V. Le.
\newblock Mnasnet: Platform-aware neural architecture search for mobile.
\newblock In {\em The IEEE Conference on Computer Vision and Pattern
  Recognition (CVPR)}, June 2019.

\bibitem{tan2019efficientnet}
Mingxing Tan and Quoc~V Le.
\newblock Efficientnet: Rethinking model scaling for convolutional neural
  networks.
\newblock {\em arXiv preprint arXiv:1905.11946}, 2019.

\bibitem{wang2019region}
Jiaqi Wang, Kai Chen, Shuo Yang, Chen~Change Loy, and Dahua Lin.
\newblock Region proposal by guided anchoring.
\newblock In {\em Proceedings of the IEEE Conference on Computer Vision and
  Pattern Recognition}, pages 2965--2974, 2019.

\bibitem{DBLP:journals/corr/abs-1906-04423}
Ning Wang, Yang Gao, Hao Chen, Peng Wang, Zhi Tian, and Chunhua Shen.
\newblock {NAS-FCOS:} fast neural architecture search for object detection.
\newblock {\em CoRR}, abs/1906.04423, 2019.

\bibitem{8354267}
P.~{Wang}, P.~{Chen}, Y.~{Yuan}, D.~{Liu}, Z.~{Huang}, X.~{Hou}, and
  G.~{Cottrell}.
\newblock Understanding convolution for semantic segmentation.
\newblock In {\em 2018 IEEE Winter Conference on Applications of Computer
  Vision (WACV)}, pages 1451--1460, March 2018.

\bibitem{williams1992simple}
Ronald~J Williams.
\newblock Simple statistical gradient-following algorithms for connectionist
  reinforcement learning.
\newblock {\em Machine learning}, 8(3-4):229--256, 1992.

\bibitem{wistuba2017finding}
Martin Wistuba.
\newblock Finding competitive network architectures within a day using uct.
\newblock {\em arXiv preprint arXiv:1712.07420}, 2017.

\bibitem{wong2018transfer}
Catherine Wong, Neil Houlsby, Yifeng Lu, and Andrea Gesmundo.
\newblock Transfer learning with neural automl.
\newblock In {\em Advances in Neural Information Processing Systems}, pages
  8356--8365, 2018.

\bibitem{xie2017aggregated}
Saining Xie, Ross Girshick, Piotr Doll{\'a}r, Zhuowen Tu, and Kaiming He.
\newblock Aggregated residual transformations for deep neural networks.
\newblock In {\em Proceedings of the IEEE conference on computer vision and
  pattern recognition}, pages 1492--1500, 2017.

\bibitem{xu2019auto}
Hang Xu, Lewei Yao, Wei Zhang, Xiaodan Liang, and Zhenguo Li.
\newblock Auto-fpn: Automatic network architecture adaptation for object
  detection beyond classification.
\newblock In {\em Proceedings of the IEEE International Conference on Computer
  Vision}, pages 6649--6658, 2019.

\bibitem{Yang2020NAS}
Antoine Yang, Pedro~M. Esperança, and Fabio~M. Carlucci.
\newblock Nas evaluation is frustratingly hard.
\newblock In {\em International Conference on Learning Representations}, 2020.

\bibitem{yao2019sm}
Lewei Yao, Hang Xu, Wei Zhang, Xiaodan Liang, and Zhenguo Li.
\newblock Sm-nas: Structural-to-modular neural architecture search for object
  detection.
\newblock {\em arXiv preprint arXiv:1911.09929}, 2019.

\bibitem{Yu_2018_ECCV}
Changqian Yu, Jingbo Wang, Chao Peng, Changxin Gao, Gang Yu, and Nong Sang.
\newblock Bisenet: Bilateral segmentation network for real-time semantic
  segmentation.
\newblock In {\em The European Conference on Computer Vision (ECCV)}, September
  2018.

\bibitem{zhu2019feature}
Chenchen Zhu, Yihui He, and Marios Savvides.
\newblock Feature selective anchor-free module for single-shot object
  detection.
\newblock In {\em Proceedings of the IEEE Conference on Computer Vision and
  Pattern Recognition}, pages 840--849, 2019.

\bibitem{zoph2016neural}
Barret Zoph and Quoc~V Le.
\newblock Neural architecture search with reinforcement learning.
\newblock {\em arXiv preprint arXiv:1611.01578}, 2016.

\bibitem{zoph2018learning}
Barret Zoph, Vijay Vasudevan, Jonathon Shlens, and Quoc~V Le.
\newblock Learning transferable architectures for scalable image recognition.
\newblock In {\em Proceedings of the IEEE conference on computer vision and
  pattern recognition}, pages 8697--8710, 2018.

\end{thebibliography}


\begin{thebibliography}{10}

\bibitem{Cordts2016Cityscapes}
Marius Cordts, Mohamed Omran, Sebastian Ramos, Timo Rehfeld, Markus Enzweiler,
  Rodrigo Benenson, Uwe Franke, Stefan Roth, and Bernt Schiele.
\newblock The cityscapes dataset for semantic urban scene understanding.
\newblock In {\em Proc. of the IEEE Conference on Computer Vision and Pattern
  Recognition (CVPR)}, 2016.

\bibitem{deng2009imagenet}
Jia Deng, Wei Dong, Richard Socher, Li-Jia Li, Kai Li, and Li~Fei-Fei.
\newblock Imagenet: A large-scale hierarchical image database.
\newblock In {\em 2009 IEEE conference on computer vision and pattern
  recognition}, pages 248--255. Ieee, 2009.

\bibitem{dong2020bench}
Xuanyi Dong and Yi~Yang.
\newblock Nas-bench-201: Extending the scope of reproducible neural
  architecture search.
\newblock {\em arXiv preprint arXiv:2001.00326}, 2020.

\bibitem{DBLP:journals/corr/abs-1811-08883}
Kaiming He, Ross~B. Girshick, and Piotr Doll{\'{a}}r.
\newblock Rethinking imagenet pre-training.
\newblock {\em CoRR}, abs/1811.08883, 2018.

\bibitem{huber1992robust}
Peter~J Huber.
\newblock Robust estimation of a location parameter.
\newblock In {\em Breakthroughs in statistics}, pages 492--518. Springer, 1992.

\bibitem{krizhevsky2009learning}
Alex Krizhevsky, Geoffrey Hinton, et~al.
\newblock Learning multiple layers of features from tiny images.
\newblock 2009.

\bibitem{lin2014microsoft}
Tsung-Yi Lin, Michael Maire, Serge Belongie, James Hays, Pietro Perona, Deva
  Ramanan, Piotr Dollar, and Larry Zitnick.
\newblock Microsoft coco: Common objects in context.
\newblock In {\em ECCV}. European Conference on Computer Vision, September
  2014.

\bibitem{schaul2015prioritized}
Tom Schaul, John Quan, Ioannis Antonoglou, and David Silver.
\newblock Prioritized experience replay.
\newblock {\em arXiv preprint arXiv:1511.05952}, 2015.

\bibitem{schulman2015high}
John Schulman, Philipp Moritz, Sergey Levine, Michael Jordan, and Pieter
  Abbeel.
\newblock High-dimensional continuous control using generalized advantage
  estimation.
\newblock {\em arXiv preprint arXiv:1506.02438}, 2015.

\bibitem{schulman2017proximal}
John Schulman, Filip Wolski, Prafulla Dhariwal, Alec Radford, and Oleg Klimov.
\newblock Proximal policy optimization algorithms.
\newblock {\em arXiv preprint arXiv:1707.06347}, 2017.

\bibitem{DBLP:journals/corr/abs-1906-04423}
Ning Wang, Yang Gao, Hao Chen, Peng Wang, Zhi Tian, and Chunhua Shen.
\newblock {NAS-FCOS:} fast neural architecture search for object detection.
\newblock {\em CoRR}, abs/1906.04423, 2019.

\bibitem{yao2019sm}
Lewei Yao, Hang Xu, Wei Zhang, Xiaodan Liang, and Zhenguo Li.
\newblock Sm-nas: Structural-to-modular neural architecture search for object
  detection.
\newblock {\em arXiv preprint arXiv:1911.09929}, 2019.

\bibitem{Yu_2018_ECCV}
Changqian Yu, Jingbo Wang, Chao Peng, Changxin Gao, Gang Yu, and Nong Sang.
\newblock Bisenet: Bilateral segmentation network for real-time semantic
  segmentation.
\newblock In {\em The European Conference on Computer Vision (ECCV)}, September
  2018.

\end{thebibliography}

\end{document}


\global\long\def\ECCVSubNumber{3177}%

\title{Appendix for CATCH: Context-based Meta Reinforcement Learning for Transferrable Architecture Search}
\titlerunning{CATCH}
\authorrunning{X. Chen and Y. Duan et al.}
\author{Xin Chen\thanks{Equal contribution.}\inst{1}, Yawen Duan$^{\star}$\inst{1}, Zewei Chen\inst{2}, Hang Xu\inst{2}, Zihao Chen\inst{2},\\ Xiaodan Liang\inst{3}, Tong Zhang\thanks{Correspondence to: tongzhang@tongzhang-ml.org}\inst{4}, Zhenguo Li\inst{2}}
\institute{The University of Hong Kong \and Huawei Noah's Ark Lab \and Sun Yat-sen University \and The Hong Kong University of Science and Technology}
\maketitle

\section{Learning Curve Comparison with Sample-based Algorithms}

\begin{figure}[H]
\begin{centering}
\includegraphics[width=0.3\columnwidth]{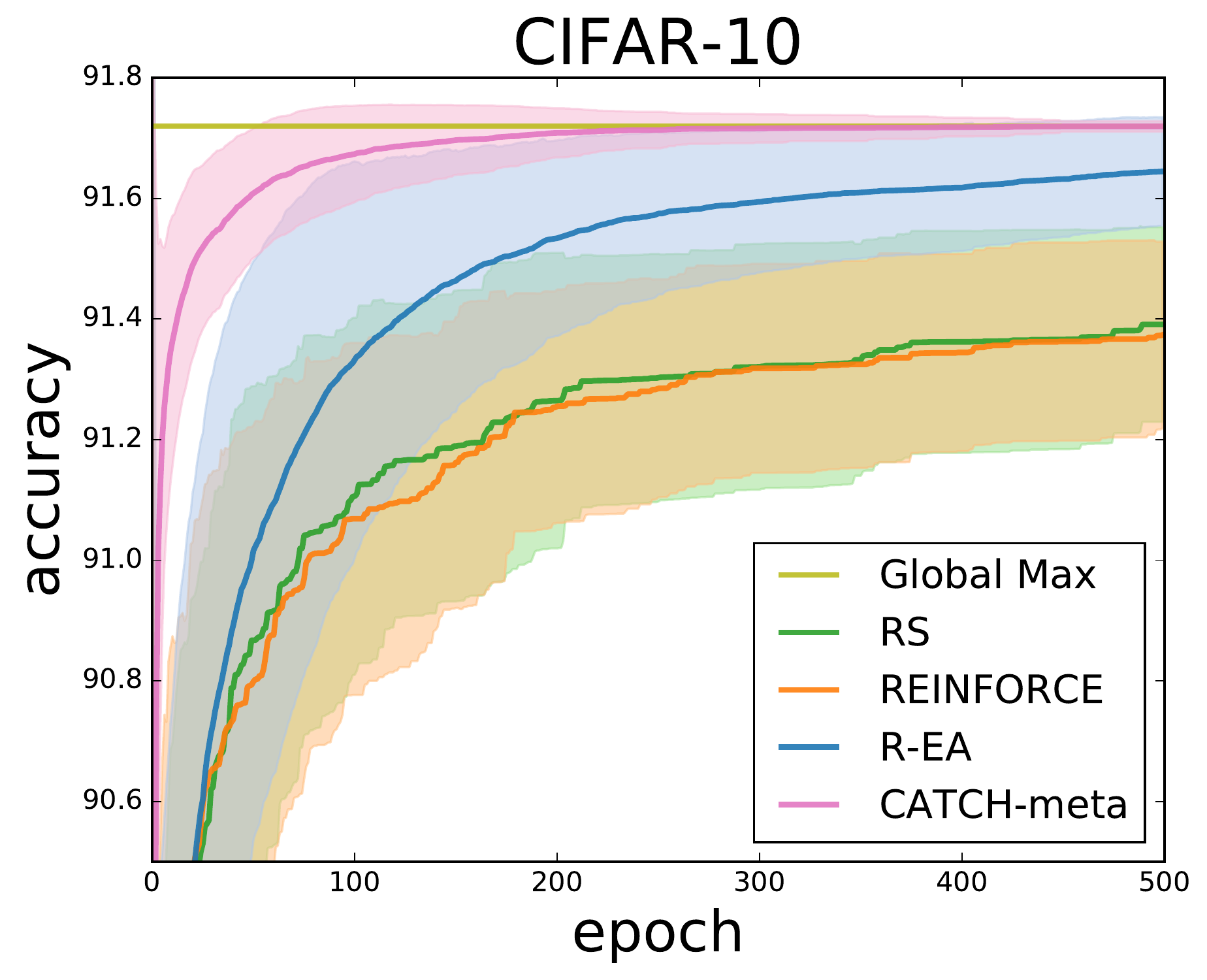}\includegraphics[width=0.3\columnwidth]{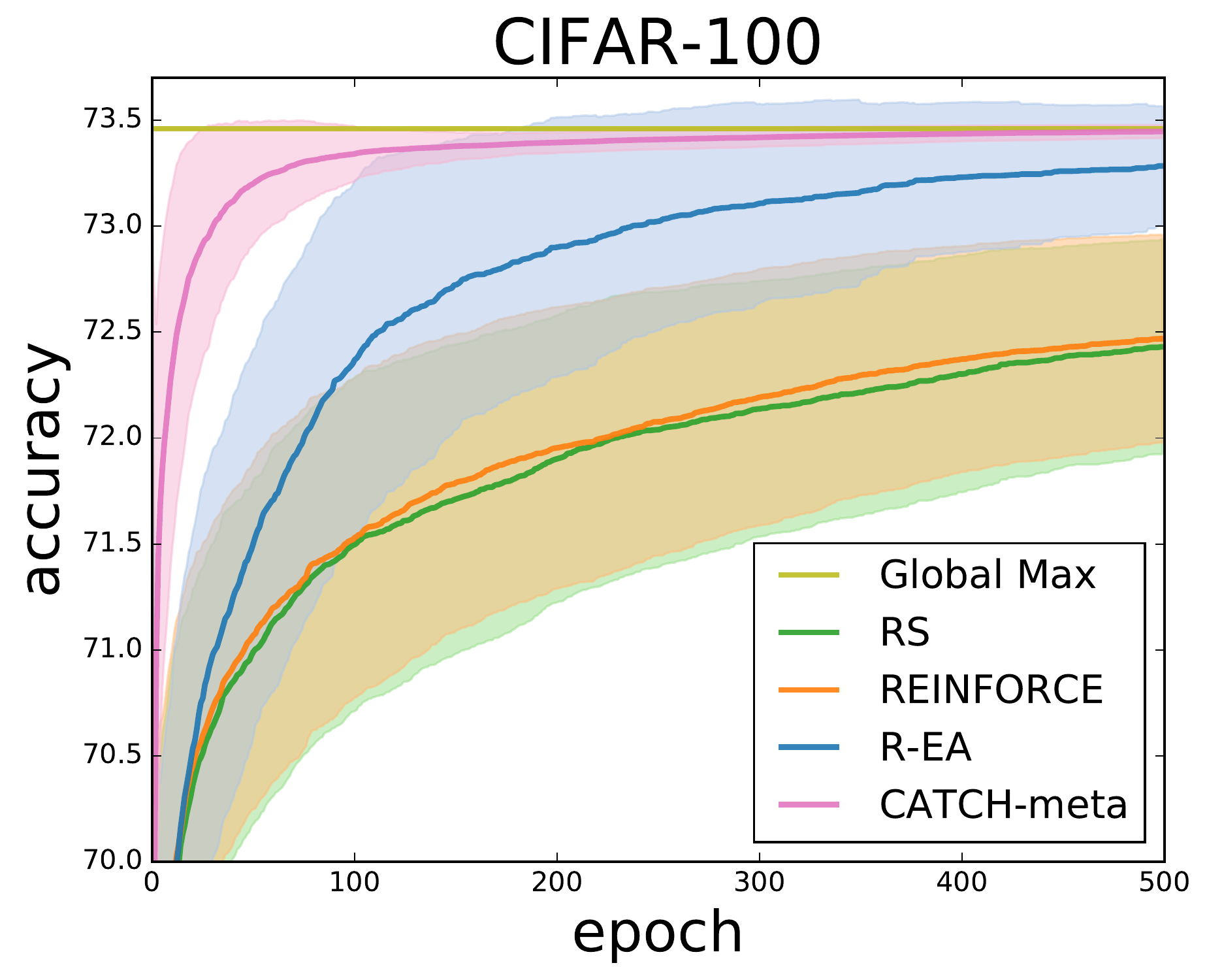}\includegraphics[width=0.3\columnwidth]{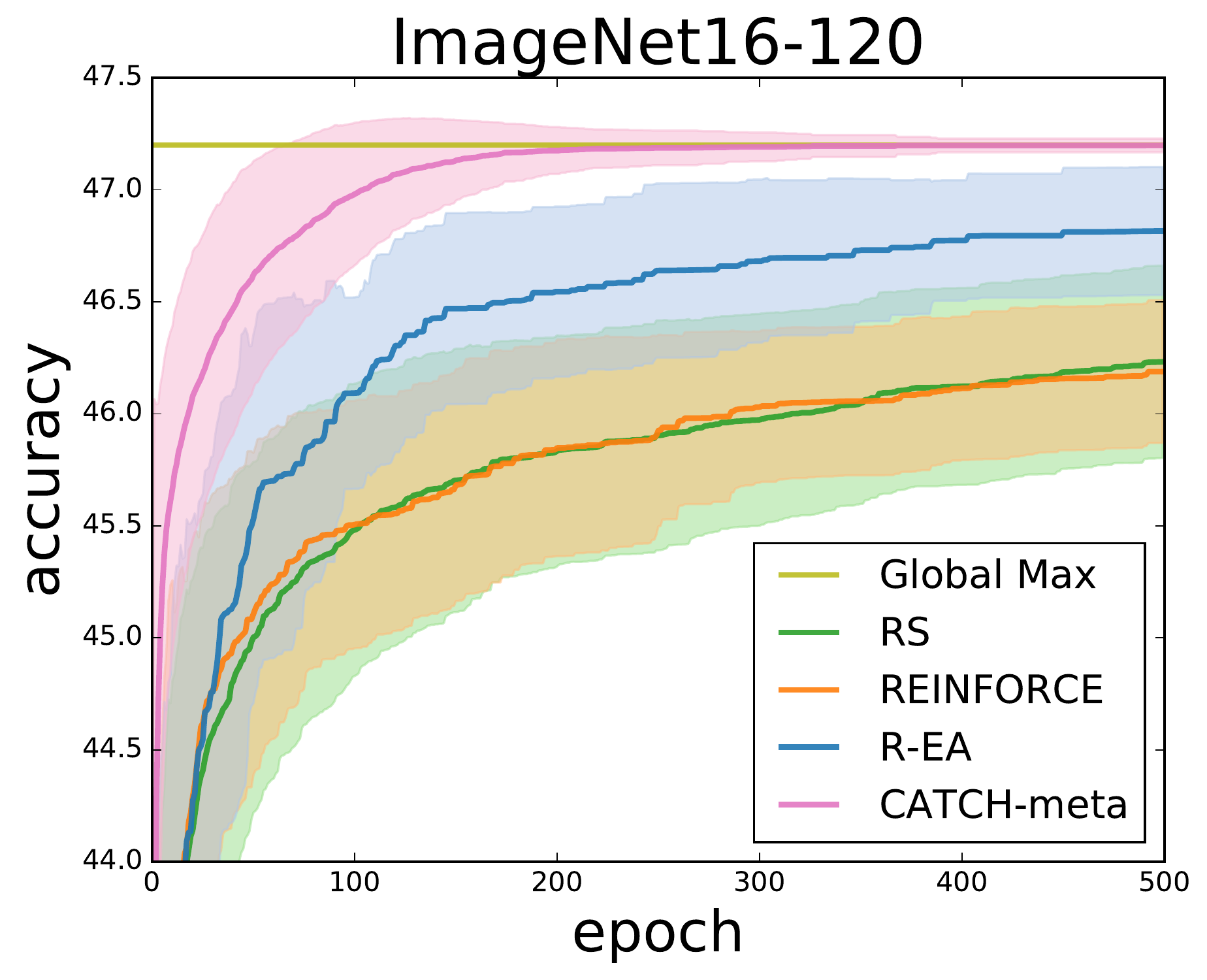}
\par\end{centering}
\caption{Comparison of CATCH with other sample-based algorithms on CIFAR-10
\cite{krizhevsky2009learning}, CIFAR-100 \cite{krizhevsky2009learning},
and ImageNet16-120 \cite{dong2020bench}.}
\end{figure}

We compare the learning curve of CATCH with other sample-based algorithms
in Figure 1. We plot each curve with the highest fully-train validation
accuracy the agent has seen at each search epoch. Each curve is plotted
with an average of 500 trials. The shaded area shows the mean $\pm$
standard deviation among all trials at each search epoch. CATCH stands
out among others with higher performance and lower variation on all
three datasets (CIFAR-10, CIFAR-100, and ImageNet16-120). It is also
on average a magnitude faster than other algorithms to find their
best architectures after 500 searching epochs. On ImageNet16-120,
none of the algorithms except CATCH could even identify the best architecture
within 500 searching epochs across all 500 trials. CATCH is also more
stable, as is indicated by its much lower variation compared with
other algorithms. Its variance tends to shrink over time, while R-EA
and REINFORCE policies are almost as unstable as random search. Through
this comparison, we further prove the adaptation speed and stability
of CATCH, along with its competency across various datasets and random
seeds. 

\section{Encoder's Adaptation Result}

\begin{figure}[t]
\begin{centering}
\includegraphics[width=0.3\columnwidth]{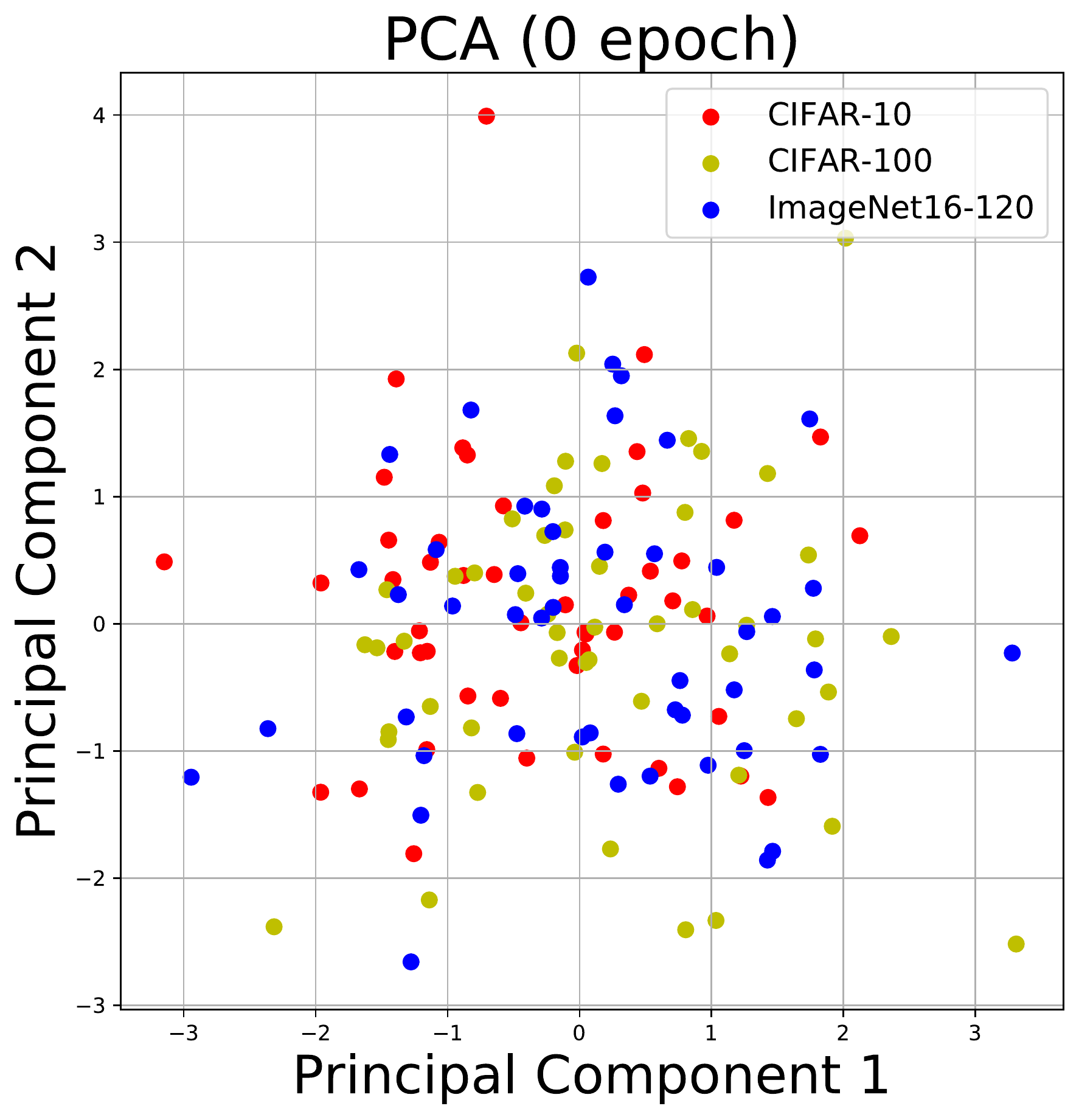}
\includegraphics[width=0.3\columnwidth]{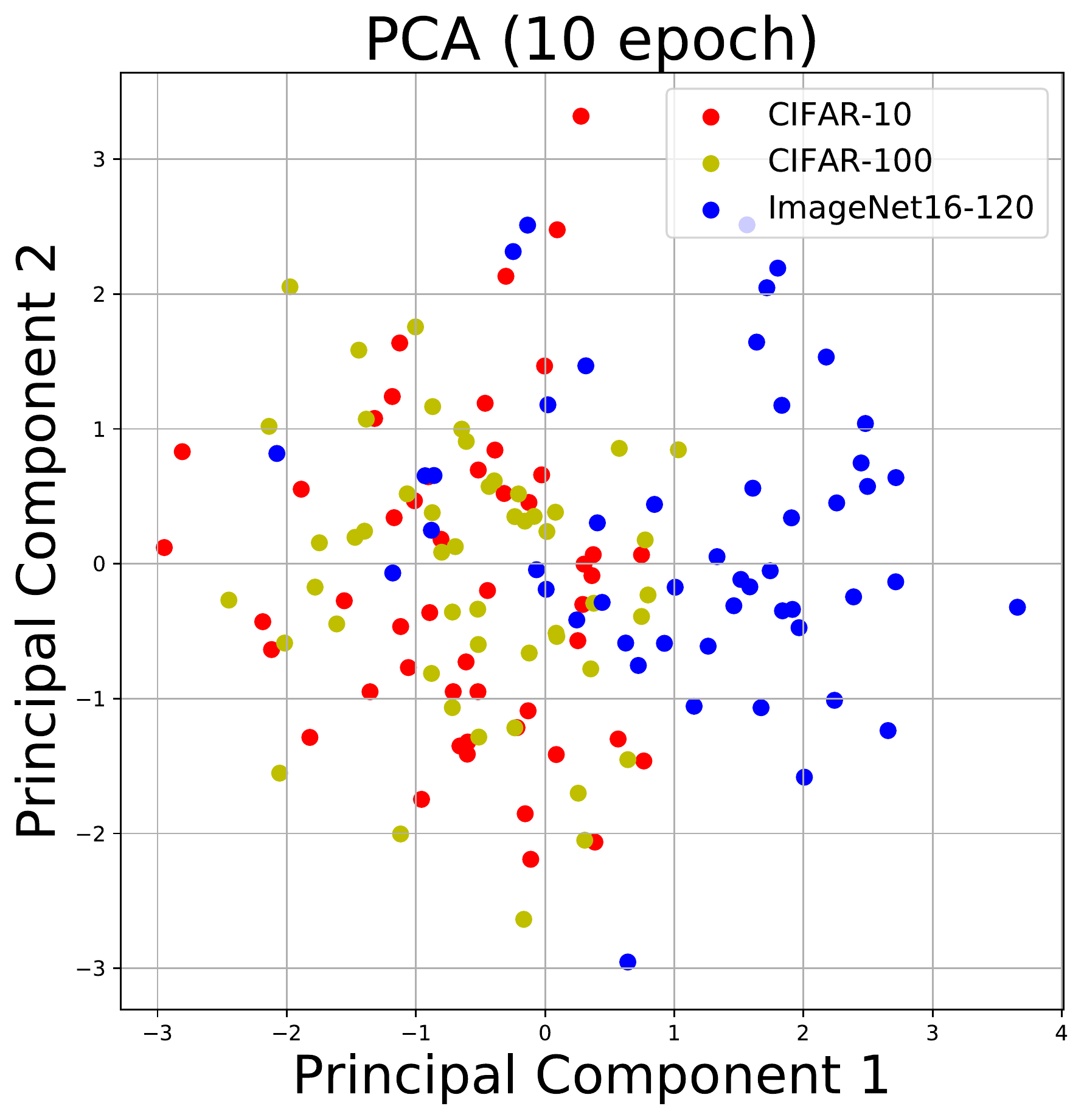}
\includegraphics[width=0.3\columnwidth]{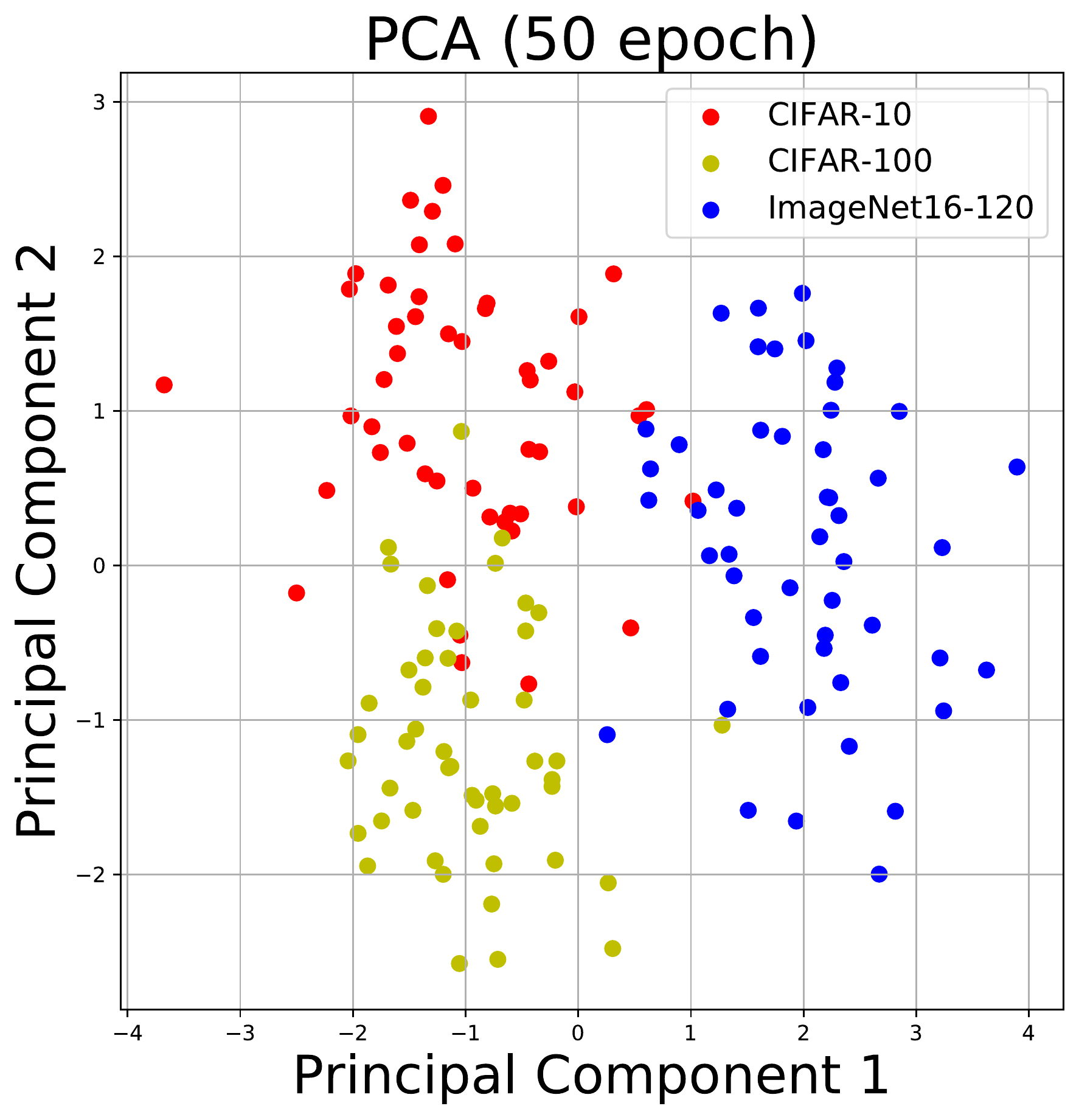}
\par\end{centering}
\caption{The encoder's adaptation process. It learns to distinguish different
datasets throughout the learning process, and thus provide informed
input to the controller and the evaluator.}
\end{figure}

Throughout the adaptation process, we hypothesize that the encoder
can provide dataset-specific guidance to the controller and the evaluator.
To test this hypothesis, we visualize the encoded latent context vector
$z$ of each dataset through Principle Component Analysis, with the
results presented in Figure 2. Each point is generated by randomly
selecting and encoding 80\% network-reward pairs from the search history.
We freeze the weights of the meta-trained controller and evaluator
policy, and only allow gradient updates for the encoder. This operation
eliminates influence from the changing controller and evaluator policies,
and thus enables us to closely observe just the behaviors of the encoder.
When the encoder is first adapted to CIFAR-10, CIFAR-100, and ImageNet16-120,
the generated context vectors are not distinguishable across the three
datasets. However, after just 10 search epochs of adaptation, we can
already identify a cluster of ImageNet16-120 context vectors. The
clusters then quickly evolve as the encoder sees more architectures.
By the 50-th search epoch, we can see three distinctive clusters as
a result of the encoder\textquoteright s fast adaptation towards the
three datasets. 

This observation is consistent with the results of NAS-Bench-201 \cite{dong2020bench}.
In the original paper, the network-performance pairs have higher correlation
between CIFAR-10 and CIFAR-100 (0.968) than that between CIFAR-10
and ImageNet16-120 (0.827). This correlation is also higher than the
correlation between CIFAR-100 and ImageNet16-120 (0.91). This attributes
to the reason why the encoder takes more search epochs to distinguish
CIFAR-10 from CIFAR-100. The results are in support of our hypothesis,
and show the encoder's capability to learn and express dataset-specific
information effectively.

\section{Ablation Study on the Evaluator}
\begin{figure}[t]
\begin{centering}
\subfloat[]{\includegraphics[scale=0.21]{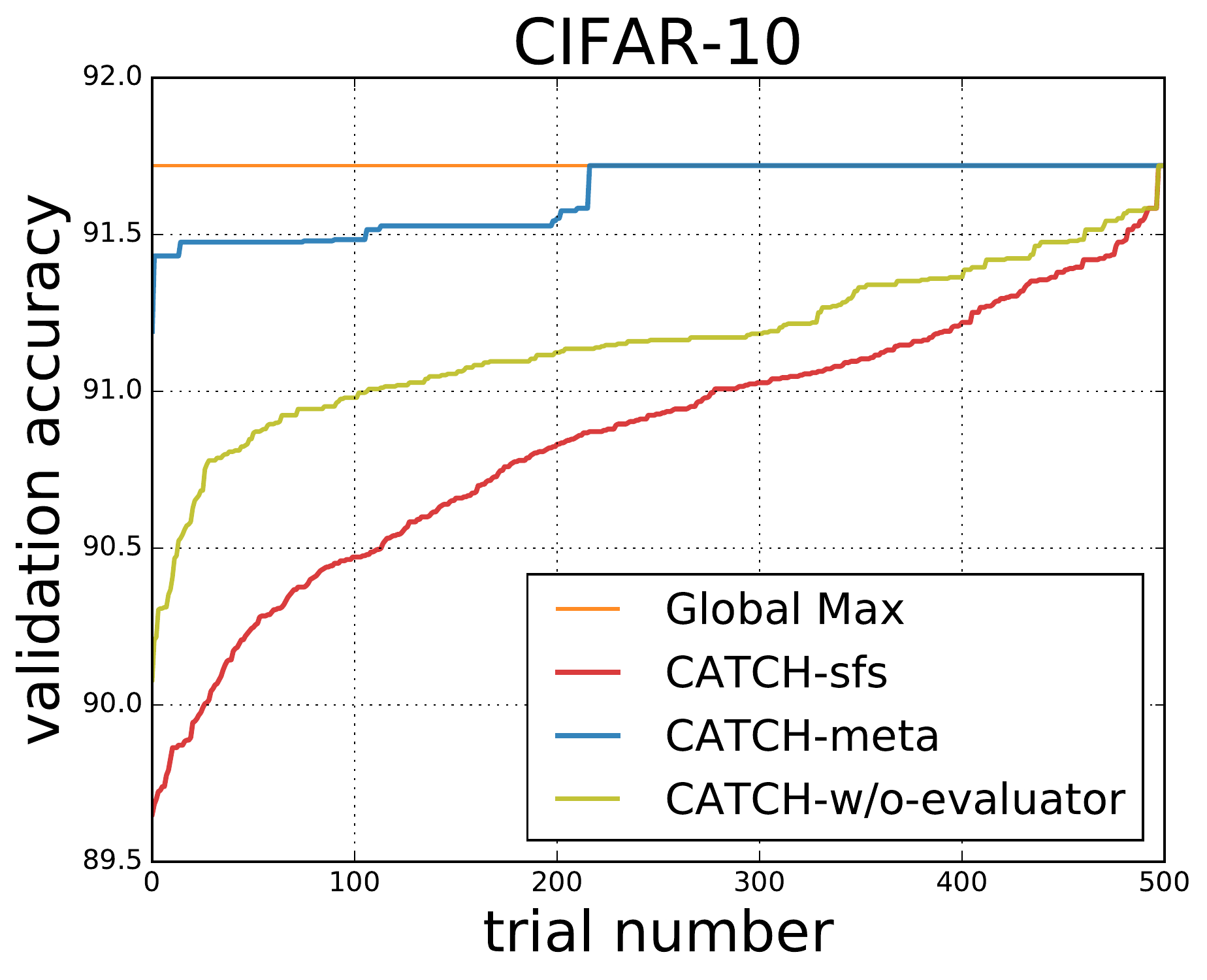}

}\subfloat[]{\includegraphics[scale=0.21]{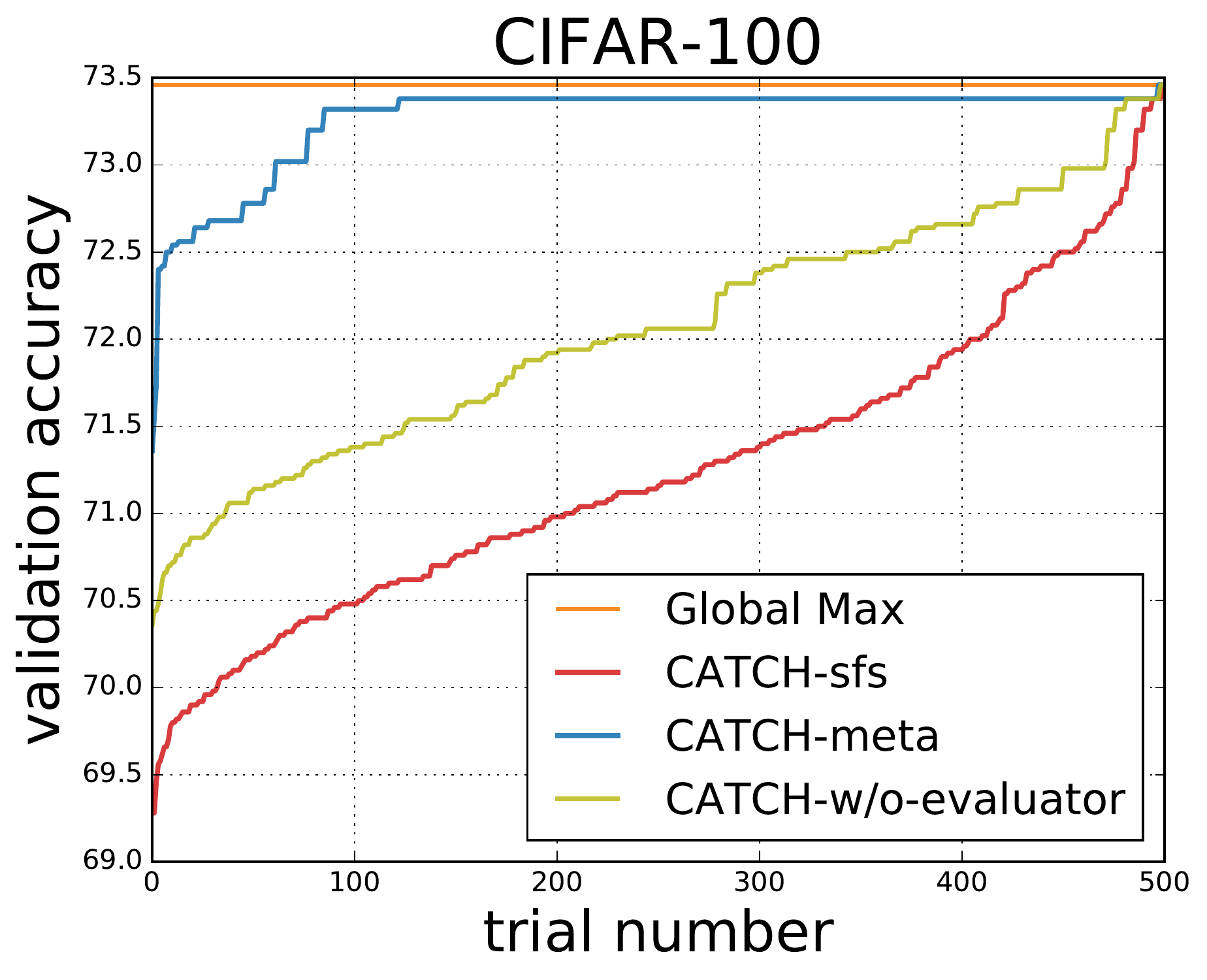}

}\subfloat[]{\includegraphics[scale=0.21]{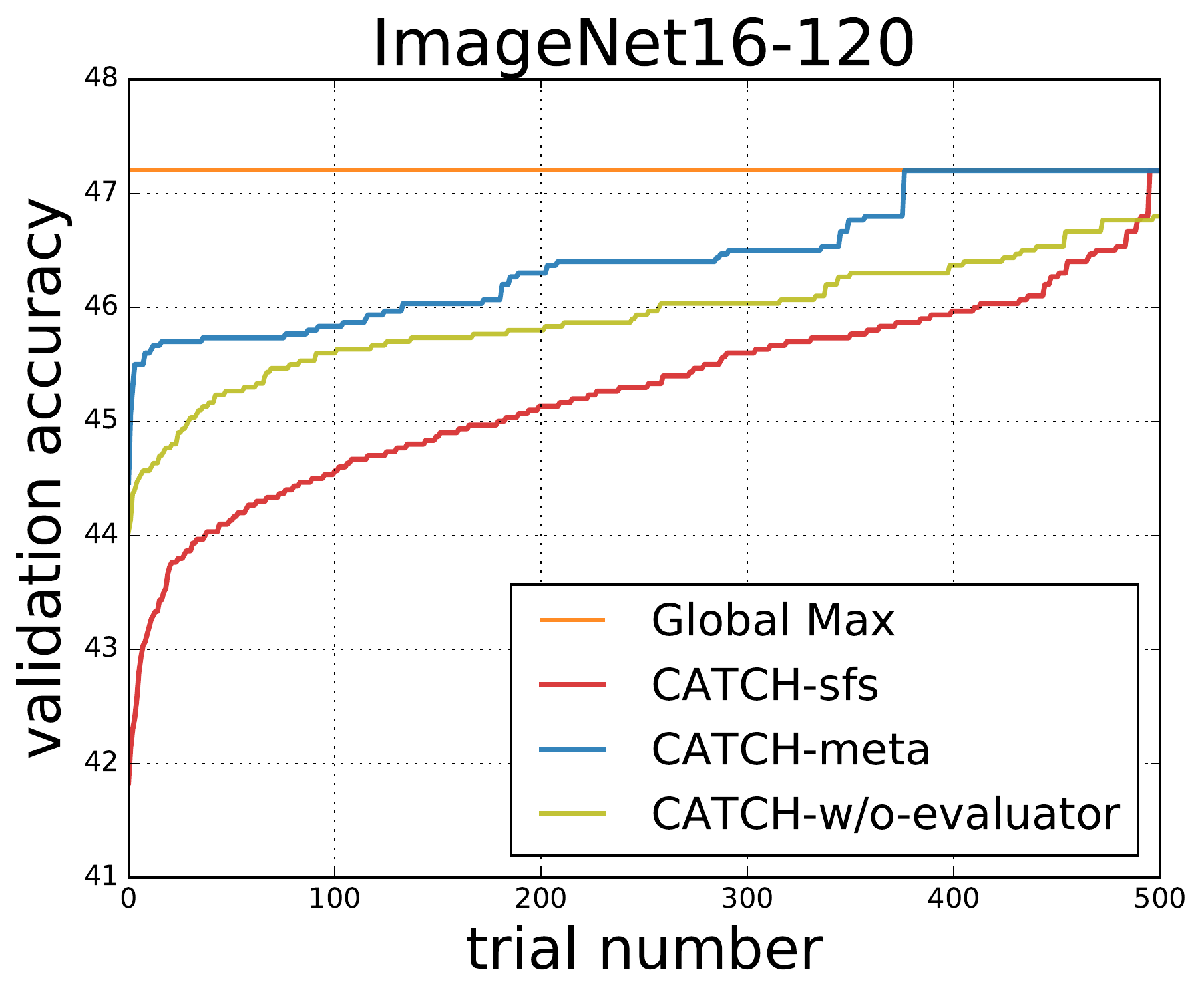}

}
\par\end{centering}
\caption{Comparison of CATCH-meta, CATCH-sfs with CATCH-without-evaluator. Including the evaluator significantly raises the performance.}

\label{t_graphs-ab-enc}
\end{figure}

We also explored the effects of the evaluator by eliminating it from
both the meta-training and adaptation phase, and its performance is
presented in Figure \ref{t_graphs-ab-enc} (a)-(c). As the figure shows,
the evaluator lifts the performance by a large margin, making it a crucial component in the search algorithm. Table \ref{t_evaluator} provides further information on the evaluator when comparing it with CATCH using ground truth as the evaluator (CATCH-GT). CATCH-GT is a hard-to-defeat baseline, but CATCH-meta managed to get very close to it and the global max accuracy.

\vspace*{-0.5 cm}

\begin{table}[]
\caption{Comparison of CATCH when using ground truth as the evaluator (CATCH-GT), CATCH without evaluator (CATCH-w/o-evaluator), and CATCH-meta. The results are taken from 100 trials where each trail contains 50 search epochs. We report the mean $\pm$ std for each setting in the table.}
\centering{}
\begin{tabular}{c|c|c|c}
\hline
                    & CIFAR-10   & CIFAR-100  & ImageNet16-120 \\ \hline
CATCH-GT            & 91.64$\pm$0.09 & 73.31$\pm$0.16 & 47.18$\pm$0.09     \tabularnewline
CATCH-w/o-evaluator & 91.17$\pm$0.25 & 72.08$\pm$0.68 & 45.86$\pm$0.54     \tabularnewline
CATCH-meta          & 91.63$\pm$0.11 & 73.29$\pm$0.31 & 46.37$\pm$0.53     \tabularnewline
Max Acc. & 91.719 & 73.45 & 47.19 \tabularnewline
\hline
\end{tabular}
\label{t_evaluator}
\end{table}

\vspace*{-0.5 cm}
\section{CATCHer Training Details}

\subsection{Controller Settings and Hyperparameters}

The controller is trained with Proximal Policy Optimization (PPO)
\cite{schulman2017proximal} algorithm, and its loss $\mathcal{L}_{c}$
is defined following the original PPO loss:

\[
\mathcal{L}_{c}=\mathbb{\hat{E}}_{t}\left[min\left(r_{t}\left(\theta_{c}\right)\hat{A_{t}},clip\left(r_{t}\left(\theta_{c}\right),1-\epsilon,1+\epsilon\right)\hat{A}_{t}\right)\right]
\]

$\epsilon$ is the PPO clipping parameter, $r_{t}\left(\theta_{c}\right)=\frac{\pi_{\theta_{c}}\left(a_{l}|s_{t}\right)}{\pi_{\theta_{old}}\left(a_{l}|s_{t}\right)}$
is the probability ratio, and $\hat{A_{t}}$ is the General Advantage
Estimate (GAE) \cite{schulman2015high} estimate:
\[
\hat{A_{t}}=\sum_{l=0}^{t}\left(\gamma\lambda\right)^{l}\delta_{l}^{V}
\]

where $\delta_{l}^{V}=r_{t}+\gamma V(s_{l+1})-V(s_{l})$ is the Bellman
residual term. The definition of $s_{l}$ can be found in Table \ref{mdp2nas}.
We show the training hyperparameters and our settings on translating
architecture search elements as Markov Decision Processes (MDP) in
the following tables. 

\begin{table}
\caption{Controller hyperparameters}

\centering{}%
\begin{tabular}{c|c|c|c}
\hline 
\multirow{2}{*}{Hyperparameter} & Value & NAS-Bench-201 \cite{dong2020bench} & Residual Block\tabularnewline
 & (meta-train) & (adaptation) & Search Space (adaptation)\tabularnewline
\hline 
Learning rate & 0.001 & 0.001 & 0.0001\tabularnewline
Adam scheduler step size & 20 & 20 & 20\tabularnewline
Adam scheduler gamma & 0.99 & 0.99 & 0.99\tabularnewline
Update frequency & 1 epoch & 1 epoch & 1 epoch\tabularnewline
Clipping parameter $\epsilon$ & 0.2 & 0.2 & 0.2\tabularnewline
Memory size & 200 & 200 & 200\tabularnewline
Discount $\gamma$ & 0.99 & 0.99 & 0.99\tabularnewline
GAE parameter $\lambda$ & 0.95 & 0.95 & 0.95\tabularnewline
Value Function coeff. & 1 & 1 & 1\tabularnewline
Entropy coeff. & 0.01 & 0.03 & 0.05\tabularnewline
\hline 
\end{tabular}
\end{table}

\begin{table}
\caption{A mapping of Neural Architecture Search elements to MDP factors for
controller training. $l$ denotes the current timestep. Invalid actions are masked by zeroing 
out their probabilities in the outputs, then softmax the remaining probabilities and sample accordingly.}

\begin{centering}
\begin{tabular}{c|c|c}
\hline 
MDP Factor & Value & Explanation\tabularnewline
\hline 
Current state $s_{l}$ & ($z,$ $[a_{1}...a_{l-1}]$) & Latent context and the current network design.\tabularnewline
Current action $a$ & $a^{l}$ & A one-hot vector of the current design choice.\tabularnewline
Reward $r$ & $R$ & A function of the evaluated network's performance.\tabularnewline
Next state $s_{l+1}$ & ($z,$ $[a_{1}...a_{l}]$) & Latent context and the current network design.\tabularnewline
\hline 
\end{tabular}
\par\end{centering}
\label{mdp2nas}
\end{table}
\vspace{-2mm}

\subsection{Encoder and Evaluator Settings}
\begin{algorithm}[t]
\begin{verbatim}
def encode_z(B, D, Contexts, Encoder):
    # Contexts: a batch of contexts {(m, r)} use for encoding
    # B: len(Contexts), batch
    # D: the dimension of latent context variable z
    # Encoder: 3-layer MLP mapping (m, r) to (mean, var) of z_i

    # encode each (m, r) to (mean, var) of z
    context_batch.rewards = normalize(context_batch.rewards)
    params = Encoder.forward(context_batch) # shape: [B, 2*D]
    
    # get mean and var; t(): matrix transpose 
    means = params[..., :D].t() # shape: [D, B]
    vars = F.softplus(params[..., D:].t()) # shape: [D, B]
    
    # get mean & var of each z_i; ds: torch.distributions
    posteriors = []
    for ms, vs in zip(unbind(means), unbind(vars)):
        z_i_mean, z_i_var = _product_of_gaussian(ms, vs)
        # form a Gaussian Posterior from z_i_mean, sqrt(z_i_var)
        z_i_posterior = ds.Gaussian(z_i_mean, sqrt(z_i_var))
        posteriors.append(z_i_posterior)
    
    # sample z from q(z|Contexts); rsample(): random sample
    z = [d.rsample() for d in posteriors]
    return torch.stack(z)
\end{verbatim}
\label{algo}
\caption{Pseudocode of Latent Context Encoding Procedure in a PyTorch-like style.}
\end{algorithm}

The encoder generates the latent conext through the network-reward information $(m, r)$. This is done by taking the encoder output as the means and variances of a $D$-dimensional Gaussian distribution, from which we sample $\boldsymbol{z}$. We provide pseudocode for this process in Algorithm 1.

\begin{table}

\caption{Encoder hyperparameters}

\begin{centering}
\begin{tabular}{c|c}
\hline 
Hyperparameter & Value\tabularnewline
\hline 
Learning rate & 0.01\tabularnewline
Dimension of $z$ & 10\tabularnewline
KL weight $\beta$ & 0.1\tabularnewline
\hline 
\end{tabular}
\par\end{centering}
\end{table}

The evaluator uses the Huber loss \cite{huber1992robust} to close the gap between its predicted network performance $\tilde{r}$ and the actual performance $r$.

\begin{equation}
\mathcal{L}_{e}=\frac{1}{n}\sum_{i}loss(r_{i},\tilde{r_{i}}),\text{where }loss(r,\tilde{r})=\begin{cases}
0.5(r_{i}-\tilde{r}_{i})^{2} & if\mid r_{i}-\tilde{r}_{i}\mid<1,\\
\mid r_{i}-\tilde{r}_{i}\mid-0.5 & otherwise.
\end{cases}
\end{equation}

\begin{table}[H]

\caption{Evaluator hyperparameters}

\begin{centering}
\begin{tabular}{c|c|c}
\hline 
\multirow{2}{*}{Hyperparameter} & Value & Value\tabularnewline
 & (meta-train) & (adaptation)\tabularnewline
\hline 
Learning rate & 0.0001 & 0.0001\tabularnewline
Exploration factor $\epsilon$ initial value & 1.0 & 0.5\tabularnewline
Exploration factor $\epsilon$ decay rate & 0.025 & 0.025\tabularnewline
Exploration factor $\epsilon$ decay step & 20 & 20\tabularnewline
Number of networks evaluated per epoch & 25 & 25\tabularnewline
PER \cite{schaul2015prioritized} prioritization factor $\alpha$ & 0.5 & 0.5\tabularnewline
PER bias correction factor $\beta$ & 0.575 & 0.575\tabularnewline
PER $\beta$ annealing step size & 0.01 & 0.01\tabularnewline
\hline 
\end{tabular}
\par\end{centering}
\end{table}

\vspace*{-0.9 cm}
\section{ImageNet, COCO, and Cityscapes Training Settings}

Table 6-8 shows our training configurations on ImageNet \cite{deng2009imagenet}
, COCO \cite{lin2014microsoft}, and Cityscapes \cite{Cordts2016Cityscapes}
. On COCO, Faster R-CNN with the ResNet backbone and Cascade FPN is
used as our baseline. It is extremely costly to perform ImageNet pretrain
for search, but training detection networks without ImageNet pretrain
was made possible by \cite{DBLP:journals/corr/abs-1811-08883}. For
COCO and Cityscapes, we use Group Normalization with halved-base-channel
groups instead of Batch Normalization. Conv2D with weight standardization
(ConvWS2D) is also applied. 

\begin{table}[H]

\caption{ImageNet training hyperparameters with 8 GPUs.}

\begin{centering}
\begin{tabular}{c|c|c}
\hline 
\multirow{2}{*}{Hyperparameter} & Value & Value\tabularnewline
 & (partial-train) & (fully-train)\tabularnewline
\hline 
Learning rate & 0.1 & 0.1\tabularnewline
Learning rate momentum & 0.9 & 0.9\tabularnewline
Weight decay & $1\times10^{-3}$ & $4\times10^{-5}$\tabularnewline
Learning rate warmup & linear for 3 epochs & linear for 3 epochs\tabularnewline
Learning rate decay policy & cosine & cosine\tabularnewline
Total epoch & 40 & 240\tabularnewline
Batch size & 1024 & 512\tabularnewline
\hline 
\end{tabular}
\par\end{centering}

\end{table}

\begin{table}[H]
\vspace{-2mm}

\caption{COCO training hyperparameters with 8 GPUs. }

\begin{centering}
\begin{tabular}{c|c|c}
\hline 
\multirow{2}{*}{Hyperparameters} & Value & Value\tabularnewline
 & (partial-train) & (fully-train)\tabularnewline
\hline 
Normalization & Group Normalization & Batch Normalization\tabularnewline
Batch size & 16 & 16\tabularnewline
Learning rate & 0.18 & 0.02\tabularnewline
Learning rate momentum & 0.9 & 0.9\tabularnewline
Weight decay & 0.0001 & 0.0001\tabularnewline
Learning rate decay policy & cosine & step\tabularnewline
Total epoch & 9 & 24\tabularnewline
\hline 
\end{tabular}
\par\end{centering}
\vspace{-2mm}

\end{table}

\begin{table}[H]
\vspace{-2mm}

\caption{Cityscapes training hyperparameters with 8 GPUs.}

\begin{centering}
\begin{tabular}{c|c|c}
\hline 
\multirow{2}{*}{Hyperparameters} & Value & Value\tabularnewline
 & (partial-train) & (fully-train)\tabularnewline
\hline 
Baseline model & BiSeNet \cite{Yu_2018_ECCV} & BiSeNet\tabularnewline
Convolution & ConvWS2D & Conv2D\tabularnewline
Normalization & Group Normalization & Synchronized BN\tabularnewline
Batch size & 32 & 16\tabularnewline
Learning rate & 0.02 & 0.025\tabularnewline
Learning rate momentum & 0.9 & 0.9\tabularnewline
Weight decay & $5\times10^{-4}$  & $1\times10^{-4}$ \tabularnewline
Learning rate warmup & linear for 5 epochs & linear for 5 epochs\tabularnewline
Learning rate decay policy & cosine & polynomial\tabularnewline
Total epoch & 40 & 100\tabularnewline
\hline 
\end{tabular}
\par\end{centering}
\vspace{-2mm}

\end{table}

\section{Searched Models of Residual Block Search Space}

We show an example model in our Residual Block search space in Figure
3. It consists of 5 stages, with depth=15, stage distribution={[}3,3,4,5{]},
and channel distribution={[}2,2,4,7{]}. We use the same notation format
to show the searched models in Table 9. 

\begin{figure}[H]
\begin{centering}
\includegraphics[height=0.26\columnwidth]{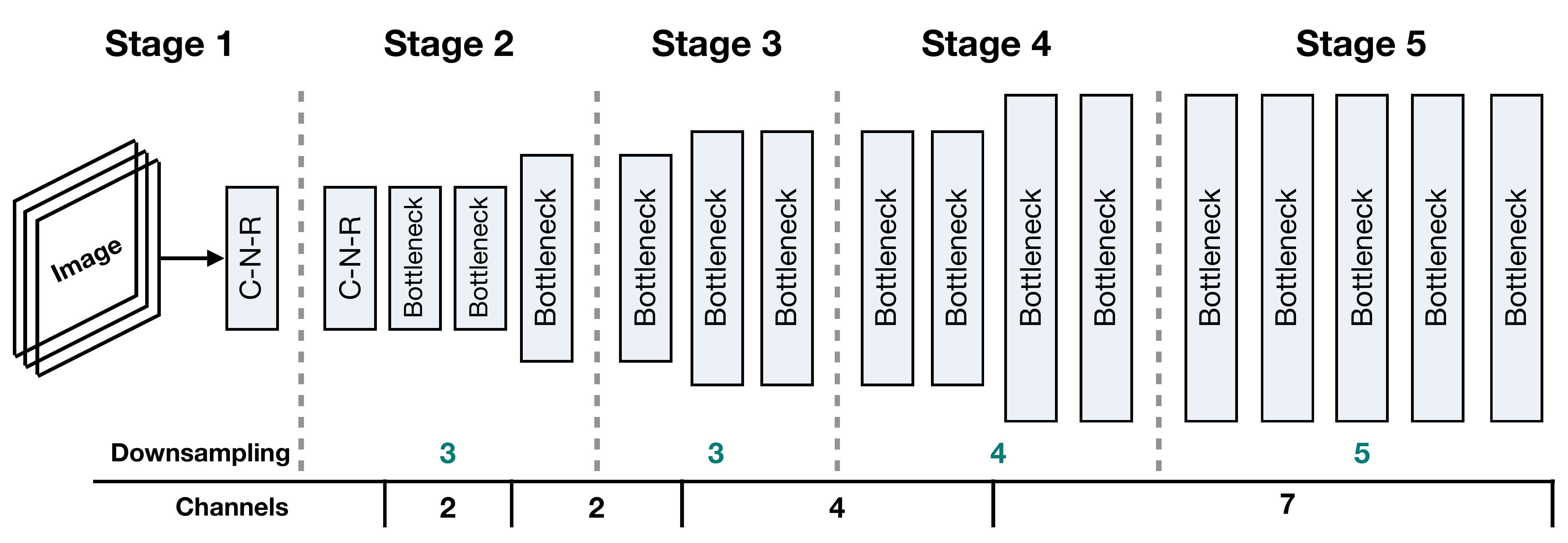}
\par\end{centering}
\caption{An example model in the Residual Block search space following \cite{yao2019sm,DBLP:journals/corr/abs-1906-04423}.
C-N-R stands for a combination of Convolution layer, Normalization
layer, and a ReLU operation.}
\end{figure}

\vspace*{-1.3 cm}

\begin{table}[H]
\caption{Searched models in Residual Block search space.}

\begin{centering}
\begin{tabular}{c|c|c|c|c|c|c}
\hline 
\multirow{2}{*}{Searched Model} & Input & \multirow{2}{*}{Depth} & Stage & Channel & \multirow{2}{*}{FLOPS(G)} & \multirow{2}{*}{Params(MB)}\tabularnewline
 & Channel &  & Distribution & Distribution &  & \tabularnewline
\hline 
CATCH-Net-A  & 64 & 20 & {[}2, 7, 8, 3{]} & {[}5, 4, 8, 3{]} & 4.45 & 25.96\tabularnewline
CATCH-Net-B  & 64 & 25 & {[}8, 5, 8, 4{]} & {[}3, 10, 8, 4{]} & 9.84 & 32.16\tabularnewline
CATCH-Net-C  & 64 & 20 & {[}5, 4, 5, 6{]} & {[}1, 8, 5, 6{]} & 8.08 & 37.03\tabularnewline
CATCH-Net-D  & 64 & 20 & {[}1, 8, 5, 6{]} & {[}2, 7, 7, 4{]} & 4.46 & 30.98\tabularnewline
\hline 
\end{tabular}
\par\end{centering}
\end{table}

\bibliographystyle{plain}
\bibliography{3177}